\documentclass[final]{article} 
\usepackage{nips13submit_e,times}
\usepackage{hyperref}
\usepackage{url}
\usepackage{graphicx}
\usepackage{amsfonts}
\usepackage{amsthm}
\usepackage{dsfont}
\usepackage{subcaption} 
\usepackage{caption}
\usepackage{enumitem}
\usepackage{algorithm}
\usepackage[noend]{algpseudocode}
\usepackage{float}
\usepackage{pgfplots}
\usepackage{epstopdf}

\newcommand{\mycomment}[1]%
{\par {\bfseries \color{green} \scriptsize{#1} \par}}

\newcommand{\quiz}[1]%
{\par {\bfseries \color{red} \scriptsize{Quiz: #1} \par}} 
\newcommand{\review}[1]%
{\par {\bfseries \color{green} \scriptsize{#1} \par}}
\newcommand{\lab}[1]%
{\par {\bfseries \color{cyan} \scriptsize{#1} \par}}
\newcommand{\concept}[1]%
{\emph{\textbf{\color{blue} #1}}}

\newcommand\cut[1]{}

\usepackage{algorithm}

\newcommand{\squishlist}{
   \begin{list}{$\bullet$}
    { \setlength{\itemsep}{0pt}      \setlength{\parsep}{3pt}
      \setlength{\topsep}{3pt}       \setlength{\partopsep}{0pt}
      \setlength{\leftmargin}{1.5em} \setlength{\labelwidth}{1em}
      \setlength{\labelsep}{0.5em} } }

\newcommand{\squishlisttwo}{
   \begin{list}{$\bullet$}
    { \setlength{\itemsep}{0pt}    \setlength{\parsep}{0pt}
      \setlength{\topsep}{0pt}     \setlength{\partopsep}{0pt}
      \setlength{\leftmargin}{2em} \setlength{\labelwidth}{1.5em}
      \setlength{\labelsep}{0.5em} } }

\newcommand{\squishend}{
    \end{list}  }
 
\newtheorem{thm}{Theorem}[section]
\newtheorem{corr}{Corollary}[section]
\newtheorem{lemma}{Lemma}[section]
\newtheorem{defn}{Definition}[section]
\newtheorem{assump}{Assumption}[section]

\newcommand{\real}{\mbox{$\mathbb{R}$}}

\newcommand{\myexpect}{\mathbb{E}}

\newcommand{\myvec}[1]{\mbox{$\mathbf{#1}$}}
\newcommand{\myvecsym}[1]{\mbox{$\boldsymbol{#1}$}}

\newcommand{\vSigma}{\mbox{$\myvecsym{\Sigma}$}}

\newcommand{\vk}{\mbox{$\myvec{k}$}}

\newcommand{\vy}{\mbox{$\myvec{y}$}}

\newcommand{\vI}{\mbox{$\myvec{I}$}}

\newcommand{\vK}{\mbox{$\myvec{K}$}}

\newcommand{\xbar}{\mbox{$\overline{x}$}}

\newcommand{\diag}{\mbox{$\mbox{diag}$}}

\newcommand{\be}{\begin{equation}}
\newcommand{\ee}{\end{equation}}
\newcommand{\bea}{\begin{eqnarray}}
\newcommand{\eea}{\end{eqnarray}}
\newcommand{\beaa}{\begin{eqnarray*}}
\newcommand{\eeaa}{\end{eqnarray*}}

\usepackage{amsmath}
\DeclareMathOperator*{\argmin}{arg\,min}
\DeclareMathOperator*{\argmax}{arg\,max}

\usepackage{algorithm}
\usepackage[noend]{algpseudocode}
\usepackage{datetime}
\usepackage{cleveref}
\usepackage{dsfont}

\crefformat{footnote}{#2\footnotemark[#1]#3}

\nipsfinalcopy
 
\title{Truncated Variance Reduction: A Unified Approach to Bayesian Optimization and Level-Set Estimation}

\newcommand{\vertiii}[1]{{\left\vert\kern-0.25ex\left\vert\kern-0.25ex\left\vert #1 \right\vert\kern-0.25ex\right\vert\kern-0.25ex\right\vert}}

\newcommand{\Mbar}{\overline{M}}

\newcommand{\cmin}{c_{\mathrm{min}}}
\newcommand{\cmax}{c_{\mathrm{max}}}
\newcommand{\gtmax}{g_{t,\mathrm{max}}}
\newcommand{\gtomax}{g_{t_{(i)},\mathrm{max}}}
\newcommand{\gtimax}{g_{t+1,\mathrm{max}}}
\newcommand{\gtsummax}{g_{t+\ell,\mathrm{max}}}
\newcommand{\deltabar}{\overline{\delta}}

\newcommand{\ALGNAME}{\textsc{TruVaR}}

\makeatletter
\newcommand\footnoteref[1]{\protected@xdef\@thefnmark{\ref{#1}}\@footnotemark}
\makeatother

\begin{document}

\maketitle

\begin{abstract}
    We present a new algorithm, truncated variance reduction (\ALGNAME), that treats Bayesian optimization (BO) and level-set estimation (LSE) with Gaussian processes in a unified fashion. The algorithm greedily shrinks a sum of truncated variances within a set of potential maximizers (BO) or unclassified points (LSE), which is updated based on confidence bounds.  \ALGNAME~is effective in several important settings that are typically non-trivial to incorporate into myopic algorithms, including pointwise costs and heteroscedastic noise.  We provide a general theoretical guarantee for \ALGNAME~covering these aspects, and use it to recover and strengthen existing results on BO and LSE.  Moreover, we provide a new result for a setting where one can select from a number of noise levels having associated costs.  We demonstrate the effectiveness of the algorithm on both synthetic and real-world data sets.
\end{abstract}

\vspace*{-2ex}
\section{Introduction}
\vspace*{-1ex}

Bayesian optimization (BO) \cite{Sha16} provides a powerful framework for automating design problems, and finds applications in robotics, environmental monitoring, and automated machine learning, just to name a few.  One seeks to find the maximum of an unknown reward function that is expensive to evaluate, based on a sequence of suitably-chosen points and noisy observations.  Numerous BO algorithms have been presented previously; see Section \ref{sec:previous_work} for an overview.  

Level-set estimation (LSE) \cite{Got13} is closely related to BO, with the added twist that instead of seeking a maximizer, one seeks to classify the domain into points that lie above or below a certain threshold.  This is of considerable interest in applications such as environmental monitoring and sensor networks, allowing one to find all ``sufficiently good'' points rather than the best point alone.  

While BO and LSE are closely related, they are typically studied in isolation.  In this paper, we provide a unified treatment of the two via a new algorithm, \emph{Truncated Variance Reduction} (\ALGNAME), which enjoys theoretical guarantees, good computational complexity, and the versatility to handle important settings such as pointwise costs, non-constant noise, and multi-task scenarios.  The main result of this paper applies to the former two settings, and even the fixed-noise and unit-cost case, we refine existing bounds via a significantly improved dependence on the noise level.

\vspace*{-1.5ex}
\subsection{Previous Work} \label{sec:previous_work}
\vspace*{-1.5ex}

Three popular myopic techniques for Bayesian optimization are expected improvement (EI), probability of improvement (PI), and Gaussian process upper confidence bound (GP-UCB) \cite{Sha16,Sri12}, each of which chooses the point maximizing an acquisition function depending directly on the current posterior mean and variance.  In \cite{Con13}, the GP-UCB-PE algorithm was presented for BO, choosing the highest-variance point within a set of potential maximizers that is updated based on confidence bounds.  Another relevant BO algorithm is BaMSOO \cite{Wan14}, which also keeps track of potential maximizers, but instead chooses points based on a global optimization technique called simultaneous online optimization (SOO).  An algorithm for level-set estimation with GPs is given in \cite{Got13}, which keeps track of a set of unclassified points.  These algorithms are computationally efficient and have various theoretical guarantees, but it is unclear how best to incorporate aspects such as pointwise costs and heteroscedastic noise \cite{Swe13}.  The same is true for the Straddle heuristic for LSE \cite{Bry08}.

Entropy search (ES) \cite{Hen12} and its predictive version \cite{Her14} choose points to reduce the uncertainty of the location of the maximum, doing so via a \emph{one-step lookahead} of the posterior rather than only the current posterior.  While this is more computationally expensive, it also permits versatility with respect to costs \cite{Swe13}, heteroscedastic noise \cite{Gol97}, and multi-task scenarios \cite{Swe13}.  A recent approach called minimum regret search (MRS) \cite{Met16} also performs a look-ahead, but instead chooses points to minimize the regret.  To our knowledge, no theoretical guarantees have been provided for these.

The multi-armed bandit (MAB) \cite{Bub12} literature has developed alongside the BO literature, with the two often bearing similar concepts.  The MAB literature is far too extensive to cover here, but we briefly mention some variants relevant to this paper.  Extensive attention has been paid to the \emph{best-arm identification} problem \cite{Jam14}, and cost constraints have been incorporated in a variety of forms \cite{Mad14}.  Moreover, the concept of ``zooming in'' to the optimal point has been explored \cite{Kle08}.  In general, the assumptions and analysis techniques in the MAB and BO literature are quite different. 

\subsection{Contributions} \label{sec:contributions}

We present a unified analysis of Bayesian optimization and level-set estimation via a new algorithm Truncated Variance Reduction (\ALGNAME).  The algorithm works by keeping track of a set of potential maximizers (BO) or unclassified points (LSE), selecting points that shrink the uncertainty within that set up to a truncation threshold, and updating the set using confidence bounds.  Similarly to ES and MRS, the algorithm performs a one-step lookahead that is highly beneficial in terms of versatility.  However, unlike these previous works, our lookahead avoids the computationally expensive task of averaging over the posterior distribution and the observations.

Also in contrast with ES and MRS, we provide theoretical bounds for \ALGNAME~characterizing the cost required to achieve a certain accuracy in finding a near-optimal point (BO) or in classifying each point in the domain (LSE).  By applying this to the standard BO setting, we not only recover existing results \cite{Got13,Con13}, but we also strengthen them via a significantly improved dependence on the noise level, with better asymptotics in the small noise limit.  Moreover, we provide a novel result for a setting in which the algorithm can choose the noise level, each coming with an associated cost.

Finally, we compare our algorithm to previous works on several synthetic and real-world data sets, observing it to perform favorably in a variety of settings.

\section{Problem Setup and Proposed Algorithm} \label{sec:algorithm}

\textbf{Setup:} 
We seek to sequentially optimize an unknown reward function $f(x)$ over a finite domain $D$.\footnote{Extensions to continuous domains are discussed in the supplementary material.}  At time $t$, we query a single point $x_t \in D$ and observe a noisy sample $y_t = f(x_t) + z_t$, where $z_t \sim N(0,\sigma^2(x_t))$ for some known noise function $\sigma^2(\cdot) \,:\, D \to \real_+$.  Thus, in general, some points may be noisier than others, in which case we have \emph{heteroscedastic noise} \cite{Gol97}.  We associate with each point a \emph{cost} according to some known cost function $c \,:\, D \to \real_+$.  If both $\sigma^2(\cdot)$ and $c(\cdot)$ are set to be constant, then we recover the standard homoscedastic and unit-cost setting.

We model $f(x)$ as a Gaussian process (GP) \cite{Ras06} having mean zero and kernel function $k(x,x')$, normalized so that $k(x,x) = 1$ for all $x \in D$. The posterior distribution of $f$ given the points and observations up to time $t$ is again a GP, with the posterior mean and variance given by \cite{Gol97}
\begin{align}
    	\mu_{t}(x) &= \vk_t(x)^T\big(\vK_t + \vSigma_t \big)^{-1} \vy_t \label{eq:mu_update} \\
    	\sigma_{t}(x)^2 &= k(x,x) - \vk_t(x)^T \big(\vK_t + \vSigma_t \big)^{-1} \vk_t(x), \label{eq:sigma_update}  
\end{align}
where $\vk_t(x) = \big[k(x_i,x)\big]_{i=1}^t$, $\vK_t = \big[k(x_t,x_{t'})\big]_{t,t'}$, and $\vSigma_t = \diag( \sigma^2(x_1), \dotsc, \sigma^2(x_t) )$.  We also let $\sigma_{t-1|x}^2(\xbar)$ denote the posterior variance of $\xbar$ upon observing $x$ along with $x_1,\cdots,x_{t-1}$.

We consider both \emph{Bayesian optimization}, which consists of finding a point whose function value is as high as possible, and \emph{level-set estimation}, which consists of classifying the domain according into points that lie above or below a given threshold $h$.  The precise performance criteria for these settings are given in Definition \ref{def:eps_acc} below.  Essentially, after spending a certain cost we report a point (BO) or a classification (LSE), but there is no preference on the values of $f(x_t)$ for the points $x_t$ chosen before coming to such a decision (in contrast with other notions such as cumulative regret).

\begin{figure}
	\centering
	\setcounter{subfigure}{0}
	\begin{subfigure}[b]{0.24\textwidth}
		\includegraphics[width=\textwidth]{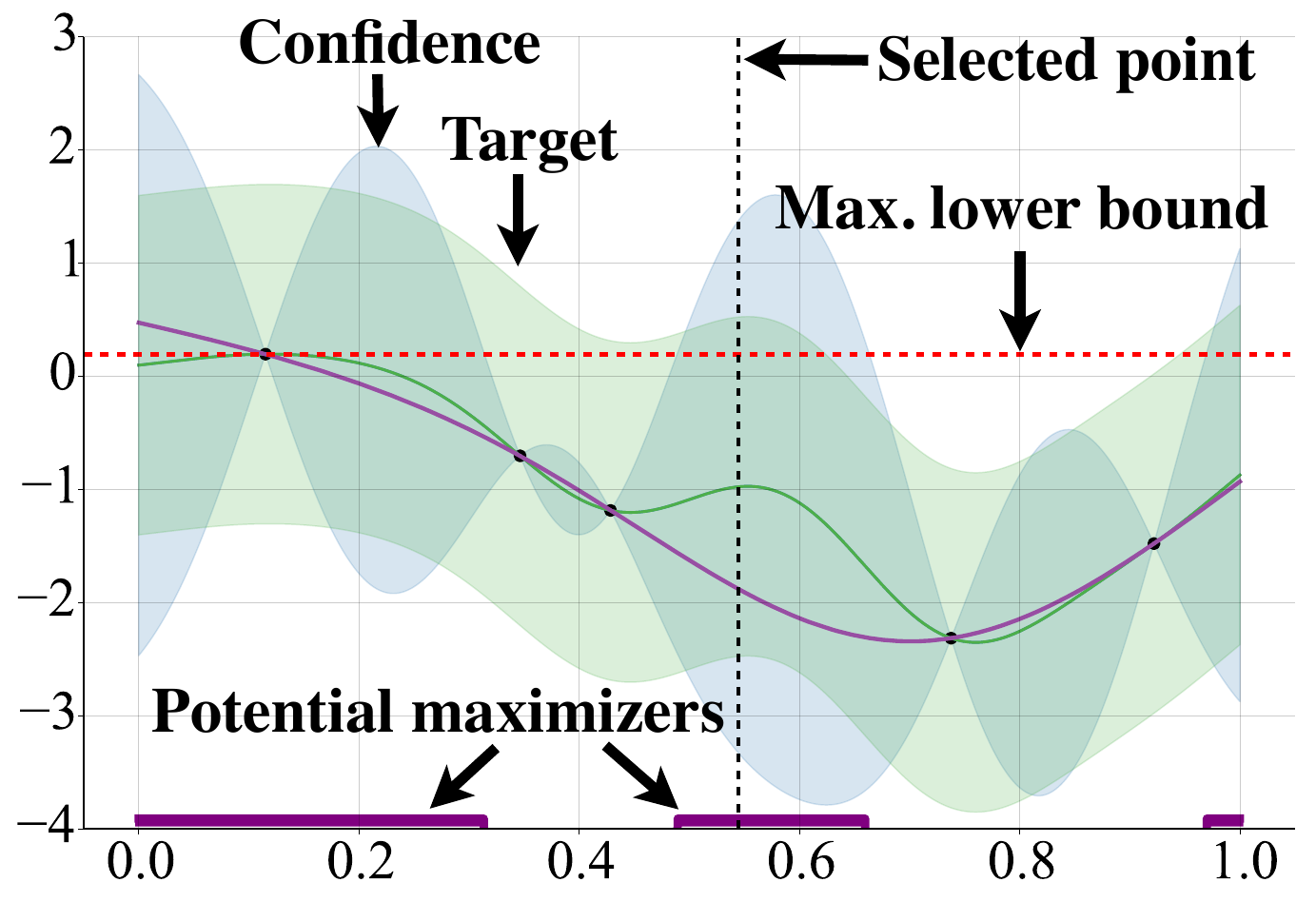}
		\caption{ $t = 6$ } 	\label{fig:}
	\end{subfigure}
	\setcounter{subfigure}{1}
	\begin{subfigure}[b]{0.24\textwidth}
		\includegraphics[width=\textwidth]{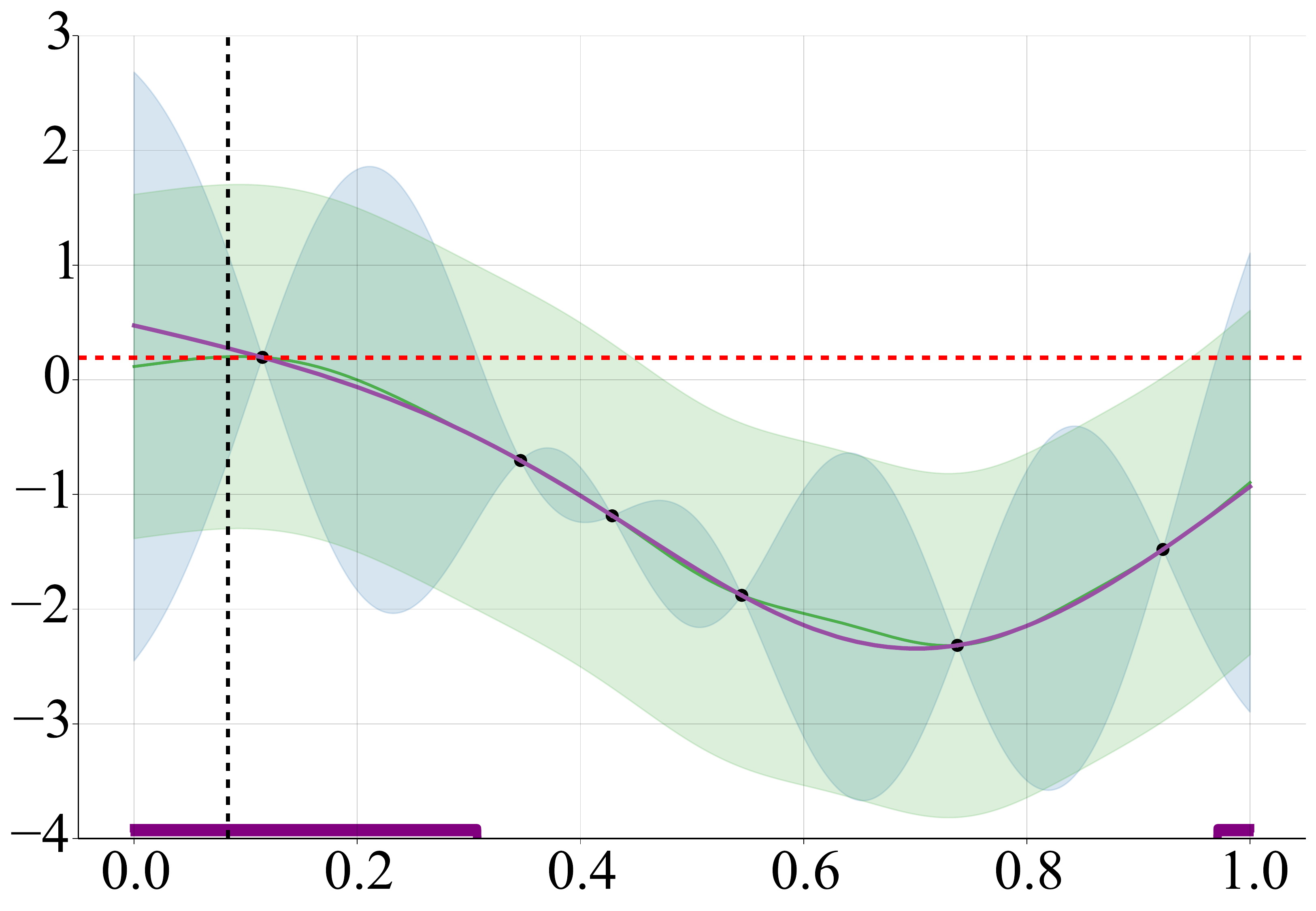}
		\caption{ $t = 7$ } \label{fig:}
	\end{subfigure}
	\setcounter{subfigure}{2}
	\begin{subfigure}[b]{0.24\textwidth}
		\includegraphics[width=\textwidth]{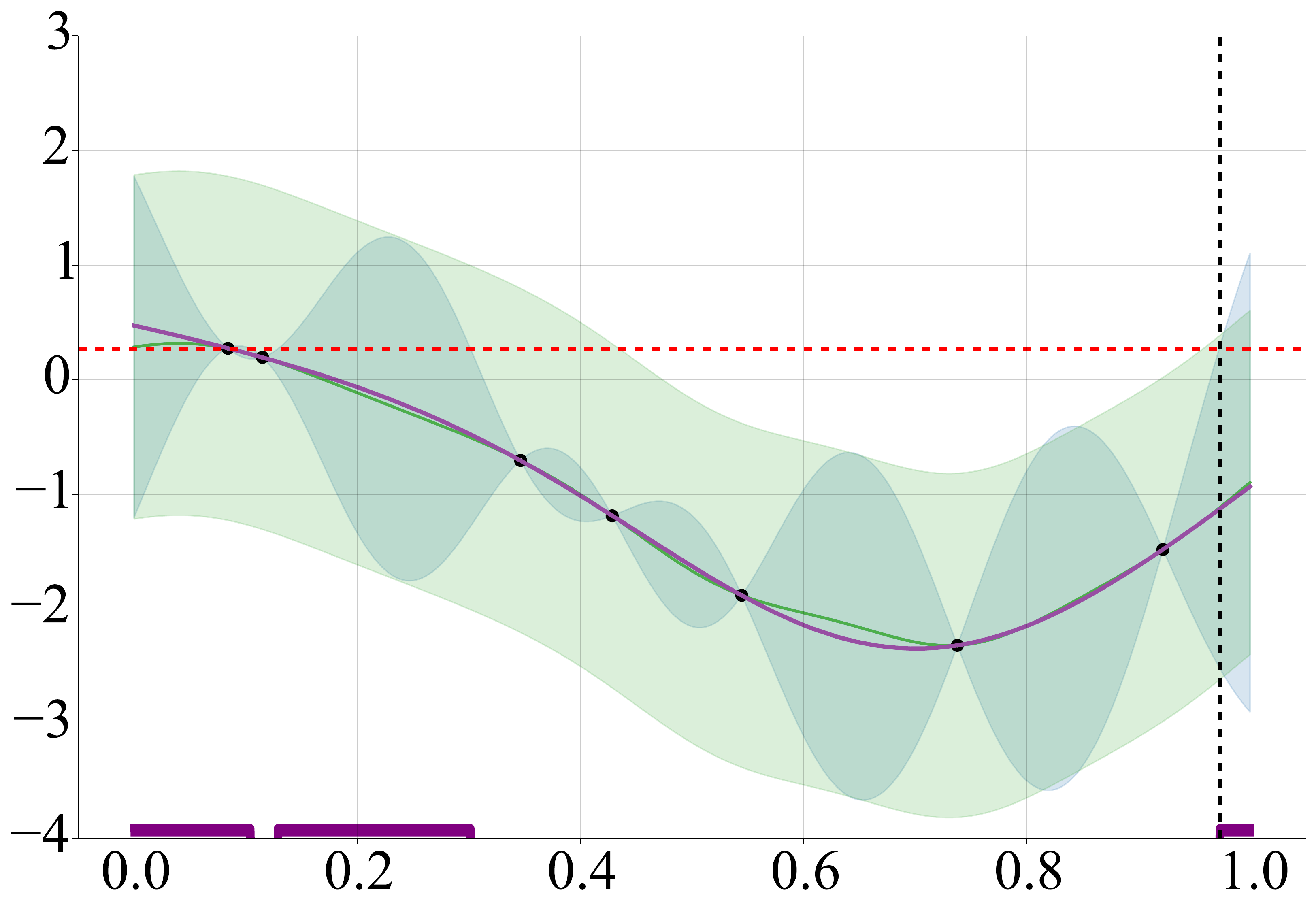}
		\caption{ $t = 8$ } \label{fig:}  
	\end{subfigure}
	\setcounter{subfigure}{3}
	\begin{subfigure}[b]{0.24\textwidth}
		\includegraphics[width=\textwidth]{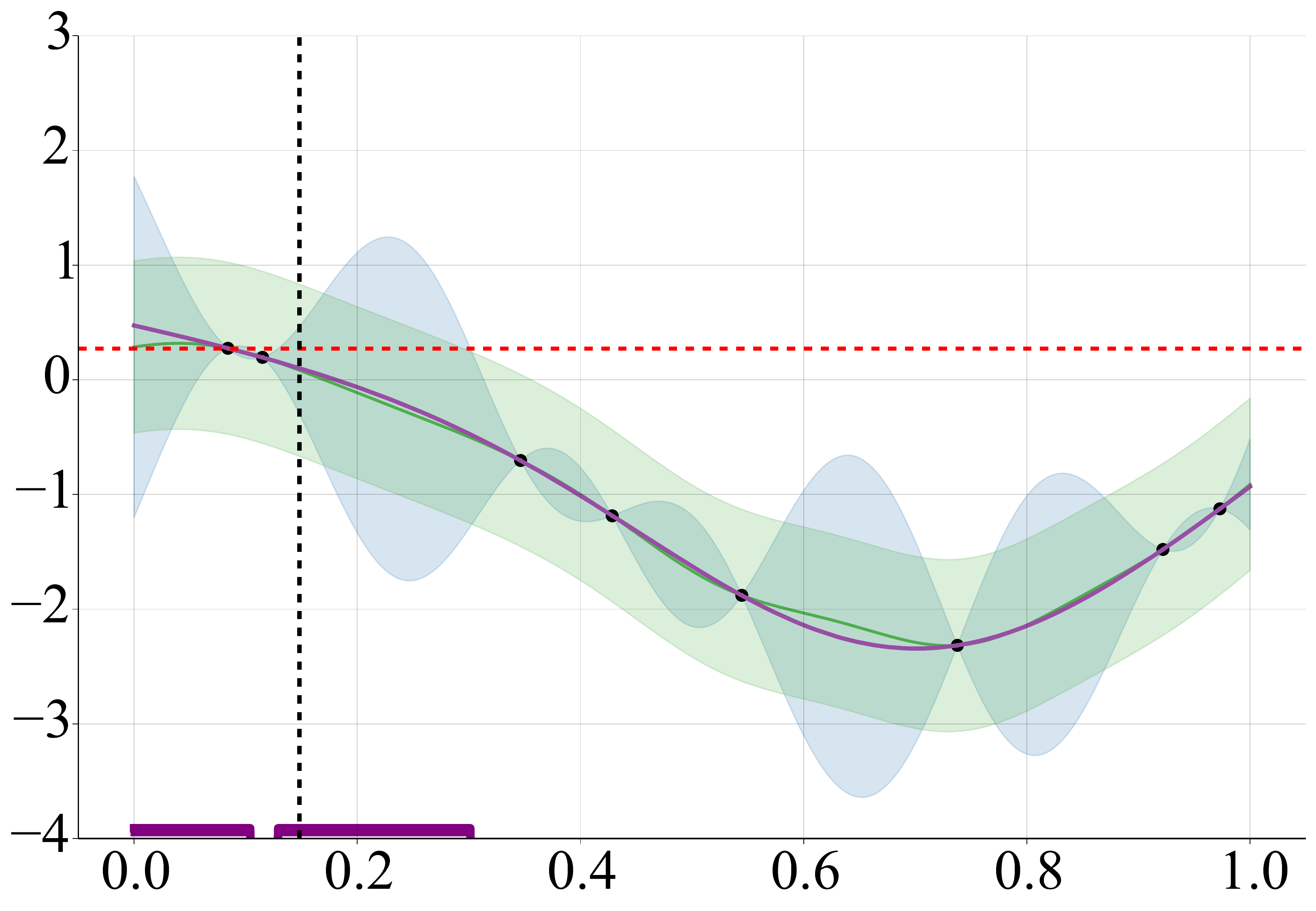}
		\caption{ $t=9$ }  
		\label{fig:}	 
	\end{subfigure}
	\caption{An illustration of the \ALGNAME~algorithm.  In (a), (b), and (c), three points within the set of potential maximizers $M_t$ are selected in order to bring the confidence bounds to within the target range, and $M_t$ shrinks during this process.  In (d), the target confidence width shrinks as a result of the last selected point bringing the confidence within $M_t$ to within the previous target.  \label{fig:toy}}
	\vspace*{-2.5ex}
\end{figure}

\textbf{\ALGNAME~algorithm:}
Our algorithm is described in Algorithm \ref{alg:truvar}, making use of the updates described in Algorithm \ref{alg:updates}. The algorithm keeps track of a sequence of unclassified points $M_t$, representing potential maximizers for BO or points close to $h$ for LSE.  This set is updated based on the confidence bounds depending on constants $\beta_{(i)}$.  The algorithm proceeds in epochs, where in the $i$-th epoch it seeks to bring the confidence $\beta_{(i)}^{1/2}\sigma_{t}(x)$ of points within $M_t$ below a target value $\eta_{(i)}$.  It does this by greedily minimizing the sum of truncated variances $\sum_{\xbar\in M_{t-1}}\max\{ \beta_{(i)} \sigma_{t-1|x}^2(\xbar), \eta_{(i)} \}$ arising from choosing the point $x$, along with a normalization and division by $c(x)$ to favor low-cost points. The truncation by $\eta_{(i)}$ in this decision rule means that once the confidence of a point is below the current target value, there is no preference in making it any lower (until the target is decreased).  Once the confidence of every point in $M_t$ is less than a factor $1+\deltabar$ above the target value, the target confidence is reduced according to a multiplication by $r \in (0,1)$.  An illustration of the process is given in Figure \ref{fig:toy}, with details in the caption.

For level-set estimation, we also keep track of the sets $H_t$ and $L_t$, containing points believed to have function values above and below $h$, respectively.  The constraint $x \in M_{t-1}$ in \eqref{eq:Mt_opt}--\eqref{eq:Lt} ensures that $\{M_t\}$ is non-increasing with respect to inclusion, and $H_t$ and $L_t$ are non-decreasing.

\begin{algorithm}
	\caption{Truncated Variance Reduction (\ALGNAME)} \label{alg:truvar}
	\begin{algorithmic}[1]
		\Require Domain $D$, GP prior ($\mu_0$, $\sigma_0$, $k$), confidence bound parameters $\deltabar > 0$, $r \in (0,1)$, $\{\beta_{(i)}\}_{i \ge 1}$, $\eta_{(1)} > 0$, and for LSE, level-set threshold $h$
		\State Initialize the epoch number $i=1$ and  potential maximizers $M_{(0)}=D$.
		\For {$t = 1,2,\dotsc$}
		\State  Choose 
			\begin{equation}
			\hspace*{-1.5ex} x_t = \argmax_{x \in D} \frac{ \sum_{\xbar\in M_{t-1}}\max\{ \beta_{(i)} \sigma_{t-1}^2(\xbar), \eta_{(i)}^2 \} - \sum_{\xbar\in M_{t-1}}\max\{ \beta_{(i)} \sigma_{t-1|x}^2(\xbar), \eta_{(i)}^2 \} }{ c(x) }. \label{eq:acq}
			\end{equation}
		\State Observe the noisy function sample $y_t$, and update according to Algorithm \ref{alg:updates} to obtain $M_t$, \hspace*{3ex}  $\mu_t$, $\sigma_t$, $l_t$ and $u_t$, as well as $H_t$ and $L_t$ in the case of LSE
		\While{$\max_{x \in M_t} \beta_{(i)}^{1/2}\sigma_{t}(x) \le (1+\deltabar) \eta_{(i)}$}
			\State Increment $i$, set $\eta_{(i)} = r \times \eta_{(i-1)}$.
		\EndWhile
		\EndFor
	\end{algorithmic}
\end{algorithm}

\begin{algorithm}
	\caption{Parameter Updates for \ALGNAME} \label{alg:updates}
	\begin{algorithmic}[1]
		\Require Selected points and observations $\{x_{t'}\}_{t'=1}^{t}$; $\{y_{t'}\}_{t'=1}^{t}$, previous sets $M_{t-1}$, $H_{t-1}$, $L_{t-1}$, parameter $\beta_{(i)}^{1/2}$, and for LSE, level-set threshold $h$.
		\State Update $\mu_t$ and $\sigma_t$ according to \eqref{eq:mu_update}--\eqref{eq:sigma_update}, and form the upper and lower confidence bounds 
		\begin{equation}
			u_t(x) = \mu_t(x) + \beta_{(i)}^{1/2}\sigma_t(x), \quad
			\ell_t(x) = \mu_t(x) - \beta_{(i)}^{1/2}\sigma_t(x). \label{eq:confidence_bounds}
		\end{equation}
		\vspace*{-2ex}
		\State For BO, set \vspace*{-2ex}
		\begin{equation}
		M_t = \bigg\{ x \in M_{t-1} \,:\, u_t(x) \ge \max_{\xbar \in M_{t-1}} \ell_t(\xbar) \bigg\}, \label{eq:Mt_opt} 		
		\end{equation}
		or for LSE, set
		\begin{gather}
			M_t = \big\{ x \in M_{t-1} \,:\, u_t(x) \ge h \text{ and } \ell_t(x) \le h \big\} \label{eq:Mt_lvl} \\
			H_t = H_{t-1} \cup \big\{ x \in M_{t-1} \,:\, \ell_t(x) > h \big\}, \quad
			L_t = L_{t-1} \cup \big\{ x \in M_{t-1} \,:\, u_t(x) < h \big\}. \label{eq:Lt} 
		\end{gather}
		\vspace*{-2ex}
	\end{algorithmic}
\end{algorithm}

The choices of $\beta_{(i)}$, $\deltabar$, and $r$ are discussed in Section \ref{sec:numerical}.  As with previous works, the kernel is assumed known in our theoretical results, whereas in practice it is typically learned from training data \cite{Sri12}.  Characterizing the effect of model mismatch or online hyperparameter updates is beyond the scope of this paper, but is an interesting direction for future work.

Some variants of our algorithm and theory are discussed in the supplementary material due to lack of space, including \emph{pure} variance reduction, non-Bayesian settings \cite{Sri12}, continuous domains \cite{Sri12}, the batch setting \cite{Con13}, and implicit thresholds for level-set estimation \cite{Got13}.

\section{Theoretical Bounds} \label{sec:theory}

In order to state our results for BO and LSE in a unified fashion, we define a notion of  \emph{$\epsilon$-accuracy} for the two settings.  That is, we define this term differently in the two scenarios, but then we provide theorems that simultaneously apply to both.  All proofs are given in the supplementary material.

\begin{defn} \label{def:eps_acc}
    After time step $t$ of \ALGNAME, we use the following terminology:
    \begin{itemize}[leftmargin=3ex, topsep=0ex] \itemsep0em 
        \item For BO, the set $M_t$ is $\epsilon$-\emph{accurate} if it contains all true maxima $x^* \in \argmax_x f(x)$, and all of its points satisfy $f(x^*) - f(x) \le \epsilon$.
        \item For LSE, the triplet $(M_t,H_t,L_t)$ is $\epsilon$-\emph{accurate} if all points in $H_t$ satisfy $f(x) > h$, all points in $L_t$ satisfy $f(x) < h$, and all points in $M_t$ satisfy $|f(x) - h| \le \frac{\epsilon}{2}$.
    \end{itemize}
    In both cases, the \emph{cumulative cost} after time $t$ is defined as $C_t = \sum_{t'=1}^t c(x_{t'})$.
\end{defn}

We use $\frac{\epsilon}{2}$ in the LSE setting instead of $\epsilon$ since this creates a region of size $\epsilon$ where the function value lies, which is consistent with the BO setting.  Our performance criterion for level-set estimation is slightly different from that of \cite{Got13}, but the two are closely related.  

\subsection{General Result}

{\bf Preliminary definitions:} Suppose that the $\{\beta_{(i)}\}$ are chosen to ensure valid confidence bounds, i.e., $l_t(x) \le f(x) \le u_t(x)$ with high probability; see Theorem \ref{thm:general} and its proof below for such choices.  In this case, we have after the $i$-th epoch that all points are either already discarded (BO) or classified (LSE), or are known up to the confidence level $(1+\deltabar)\eta_{(i)}$.  For the points with such confidence, we have $u_t(x) - l_t(x) \le 2(1+\deltabar) \eta_{(i)}$, and hence 
\begin{equation}
    u_t(x) \le l_t(x) + 2(1+\deltabar)\eta_{(i)} \le f(x) + 2(1+\deltabar)\eta_{(i)},
\end{equation}
and similarly $l_t(x) \ge f(x) - 2(1+\deltabar)\eta_{(i)}$.  This means that all points other than those within a gap of width $4(1+\deltabar)\eta_{(i)}$ must have been discarded or classified:
\begin{align}
    M_t &\subseteq \big\{ x \,:\, f(x) \ge f(x^*) - 4(1+\deltabar)\eta_{(i)} \big\} =: \Mbar_{(i)} \quad \text{(BO)} \label{eq:Mbar_opt} \\
    M_t &\subseteq \big\{ x \,:\, |f(x) - h| \le 2(1+\deltabar)\eta_{(i)} \big\} =: \Mbar_{(i)} \quad \text{(LSE)} \label{eq:Mbar_lvl}
\end{align}
Since no points are discarded or classified initially, we define $\Mbar_{(0)} = D$.

For a collection of points $S = (x'_1,\dotsc,x'_{|S|})$, possibly containing duplicates, we write the total cost as $c(S) = \sum_{i=1}^{|S|} c(x'_i)$.  Moreover, we denote the posterior variance upon observing the  points up to time $t-1$ \emph{and} the additional points in $S$ by $\sigma_{t-1|S}(\xbar)$.  Therefore, $c(x) = c(\{x\})$ and $\sigma_{t-1|x}(\xbar) = \sigma_{t-1|\{x\}}(\xbar)$.   The minimum cost (respectively, maximum cost) is denoted by $\cmin = \min_{x \in D} c(x)$ (respectively, $\cmax = \max_{x \in D} c(x)$).

Finally, we introduce the quantity
\begin{equation}
    C^*(\xi,M) = \min_S\Big\{ c(S) \,:\, \max_{\xbar \in M}\sigma_{0|S}(\xbar) \le \xi \Big\}, \label{eq:C*}
\end{equation}
representing the minimum cost to achieve a posterior standard deviation of at most $\xi$ within $M$.

{\bf Main result:} In all of our results, we make the following assumption.

\begin{assump}
    The kernel $k(x,x')$ is such that the variance reduction function
    \begin{equation}
        \psi_{t,x}(S) = \sigma^2_t(x) - \sigma^2_{t|S}(x)
    \end{equation}
    is submodular \cite{Kra12} for any time $t$, and any selected points $(x_1,\dotsc,x_t)$ and query point $x$.
\end{assump} 

This assumption has been used in several previous works based on Gaussian processes, and sufficient conditions for its validity can be found in \cite[Sec.~8]{Das08}.   We now state the following general guarantee.

\begin{thm} \label{thm:general}
    Fix $\epsilon >0$ and $\delta \in (0,1)$, and suppose there exist values $\{C_{(i)}\}$ and $\{\beta_{(i)}\}$ such that
    \begin{equation}
        C_{(i)} \ge C^*\bigg(\frac{\eta_{(i)}}{\beta_{(i)}^{1/2}}, \Mbar_{(i-1)}\bigg) \log\frac{ |\Mbar_{(i-1)}| \beta_{(i)} }{ \deltabar^2 \eta_{(i)}^2 } + \cmax, \label{eq:Ci_bound}
    \end{equation}
    and
    \begin{equation}
        \beta_{(i)} \ge 2\log\frac{ |D| \big(\sum_{i' \le i} C_{(i')}\big)^2 \pi^2 }{ 6\delta \cmin^2 }. \label{eq:beta_i}
   \end{equation}
    Then if \ALGNAME~is run with these choices of $\beta_{(i)}$ until the cumulative cost reaches
    \begin{equation}
        C_{\epsilon} = \sum_{i \,:\, 4(1+\deltabar)\eta_{(i-1)} > \epsilon} C_{(i)}, \label{eq:C_eps}
    \end{equation}
    then with probability at least $1-\delta$, we have $\epsilon$-accuracy.
\end{thm}

While this theorem is somewhat abstract, it captures the fact that the algorithm improves when points having a lower cost and/or lower noise are available, since both of these lead to a smaller value of $C^*(\xi,M)$; the former by directly incurring a smaller cost, and the latter by shrinking the variance more rapidly.  Below, we apply this result to some important cases.

\subsection{Results for Specific Settings}

{\bf Homoscedastic and unit-cost setting:} Define the maximum mutual information \cite{Sri12}
\begin{equation} 
    \gamma_T = \max_{x_1,\dotsc,x_T} \frac{1}{2}\log\det\big( \vI_T + \sigma^{-2} \vK_T \big), \label{eq:gamma_def}
\end{equation}
and consider the case that $\sigma^2(x) = \sigma^2$ and $c(x) = 1$. In the supplementary material, we provide a theorem with a condition for $\epsilon$-accuracy of the form $T \ge \Omega^*\big( \frac{C_1 \gamma_T \beta_T}{ \epsilon^2 } + 1\big)$ with $C_1 = \frac{1}{\log(1+\sigma^{-2})}$, thus matching \cite{Got13,Con13} up to logarithmic factors.  In the following, we present a refined version that has a significantly better dependence on the noise level, thus exemplifying that a more careful analysis of \eqref{eq:Ci_bound} can provide improvements over the standard bounding techniques. 

\begin{corr} \label{thm:noise_dep}
    Fix $\epsilon >0$ and $\delta \in (0,1)$, define $\beta_T = 2\log\frac{|D| T^2 \pi^2}{ 6\delta }$, and set $\eta_{(1)} = 1$ and $r = \frac{1}{2}$.  There exist choices of $\beta_{(i)}$ (not depending on the time horizon $T$) such that we have $\epsilon$-accuracy with probability at least $1-\delta$ once the following condition holds:
    \begin{align}
        T &\ge  \bigg( 2\sigma^2 \gamma_T \beta_T \frac{ 96(1+\deltabar)^2  }{ \epsilon^2 } + C_1 \gamma_T \beta_T \frac{ 6(1+\deltabar)^2 }{ \sigma^2 } + 2\Big\lceil \log_{2}\frac{32(1+\deltabar)^2 }{\epsilon\sigma} \Big\rceil \bigg) \log\frac{ 16(1+\deltabar)^2 |D| \beta_T }{ \deltabar^2 \epsilon^2 }, \label{eq:improved}
    \end{align}
    where $C_1 = \frac{1}{\log(1+\sigma^{-2})}$.  This condition is of the form $T \ge \Omega^*\big( \frac{\sigma^2\gamma_T \beta_T}{ \epsilon^2 } + \frac{C_1\gamma_T\beta_T}{\sigma^2} + 1\big)$.
\end{corr}

The choices  $\eta_{(1)} = 1$ and $r = \frac{1}{2}$ are made for mathematical convenience, and a similar result follows for any other choices $\eta_{(1)} > 0$ and $r\in(0,1)$, possibly with different constant factors.

As $\sigma^2 \to \infty$ (i.e., high noise), both of the above-mentioned bounds have noise dependence $O^*(\sigma^2)$, since $\log(1+\alpha^{-1}) = O(\alpha^{-1})$ as $\alpha\to\infty$.  On the other hand, as $\sigma^2 \to 0$ (i.e., low noise), $C_1$ is logarithmic, and Corollary \ref{thm:noise_dep} is significantly better provided that $\epsilon \ll \sigma$.  

{\bf Choosing the noise and cost:} Here we consider the setting that there is a domain of points $D_0$ that the reward function depends on, and alongside each point we can \emph{choose} a noise variance $\sigma^2(k)$ ($k=1,\dotsc,K$).  Hence, $D = D_0 \times \{1,\cdots,K\}$.  Lower noise variances incur a higher cost according to a cost function $c(k)$.

 \begin{corr} \label{thm:choose_noise}
    For each $k=1,\cdots,K$, let $T^*(k)$ denote the smallest value of $T$ such that \eqref{eq:improved} holds with $\sigma^2(k)$ in place of $\sigma^2$, and with $\beta_T = 2\log\frac{|D| T^2 \cmax^2 \pi^2}{ 6\delta\cmin^2 }$.  Then, under the preceding setting, there exist choices of $\beta_{(i)}$ (not depending on $T$) such that we have $\epsilon$-accuracy with probability at least $1-\delta$ once the cumulative cost reaches $\min_k c(k) T^*(k)$.
 \end{corr}
 
This result roughly states that we obtain a bound as good as that obtained by sticking to \emph{any fixed choice of noise level}.  In other words, every choice of noise (and corresponding cost) corresponds to a different version of a BO or LSE algorithm (e.g., \cite{Got13,Con13}), and our algorithm has a similar performance guarantee to the best among all of those.  This is potentially useful in avoiding the need for running an algorithm once per noise level and then choosing the best-performing one.  Moreover, we found numerically that beyond matching the best fixed noise strategy, we can strictly improve over it by mixing the noise levels; see Section \ref{sec:numerical}.

\vspace*{-2ex}
\section{Experimental Results} \label{sec:numerical}
\vspace*{-1.5ex}

We evaluate our algorithm in both the level-set estimation and Bayesian optimization settings. 

\textbf{Parameter choices:} As with previous GP-based algorithms that use confidence bounds, our theoretical choice of $\beta_{(i)}$ in \ALGNAME~is typically overly conservative.  Therefore, instead of using \eqref{eq:beta_i} directly, we use a more aggressive variant with similar dependence on the domain size and time: $\beta_{(i)} = a \log(|D|t_{(i)}^2)$, where $t_{(i)}$ is the time at which the epoch starts, and $a$ is a constant.  Instead of the choice $a=2$ dictated by \eqref{eq:beta_i}, we set $a = 0.5$ for BO to avoid over-exploration.  We found exploration to be slightly more beneficial for LSE, and hence set $a=1$ for this setting.  We found \ALGNAME~to be quite robust with respect to the choices of the remaining parameters, and simply set $\eta_{(1)} = 1$, $r=0.1$, and $\deltabar = 0$ in all experiments; while our theory assumes $\deltabar > 0$, in practice there is negligible difference between choosing zero and a small positive value.

\textbf{Level-set estimation:} 
For the LSE experiments, we use a common classification rule in all algorithms, classifying the points according to the posterior mean as $\hat{H}_t = \lbrace x:  \mu_t(x) \geq h \rbrace$ and $\hat{L}_t = \lbrace x: \mu_t(x) < h \rbrace$.  The classification accuracy is measured by the $F_1$-score (i.e., the harmonic mean of precision and recall) with respect to the true super- and sub-level sets.

We compare \ALGNAME~against the GP-based LSE algorithm \cite{Got13}, which we name via the authors' surnames as GCHK, as well as the state-of-the-art straddle (STR) heuristic~\cite{Bry08} and the maximum variance rule (VAR) \cite{Got13}.  Descriptions can be found in the supplementary material.  GCHK includes an exploration constant $\beta_t$, and we follow the recommendation in \cite{Got13} of setting $\beta_t^{1/2} = 3$. 

\begin{figure}
	\centering
	\setcounter{subfigure}{0}
	\begin{subfigure}[b]{0.32\textwidth}
		\includegraphics[width=\textwidth]{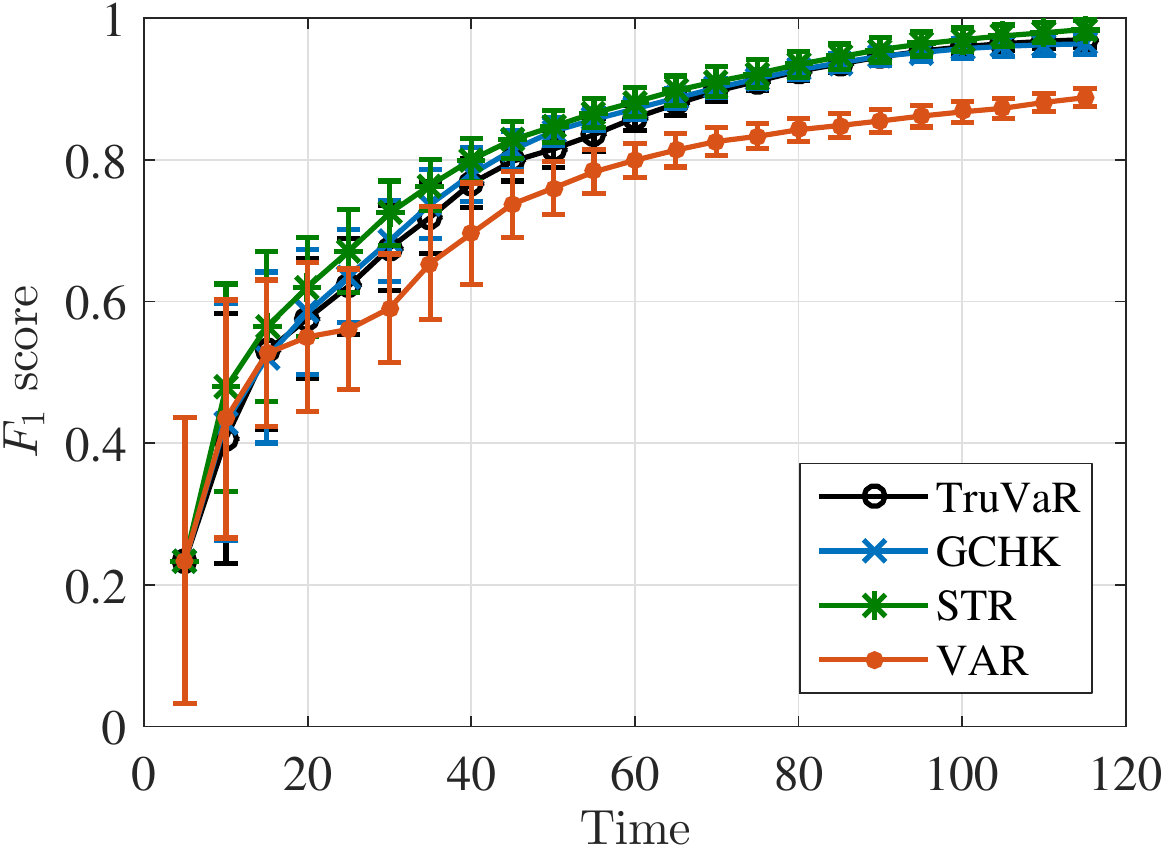}
		\caption{Lake data, unit-cost} \label{fig:lake_time}
	\end{subfigure}
	\setcounter{subfigure}{1}
	\begin{subfigure}[b]{0.31\textwidth}
		\includegraphics[width=\textwidth]{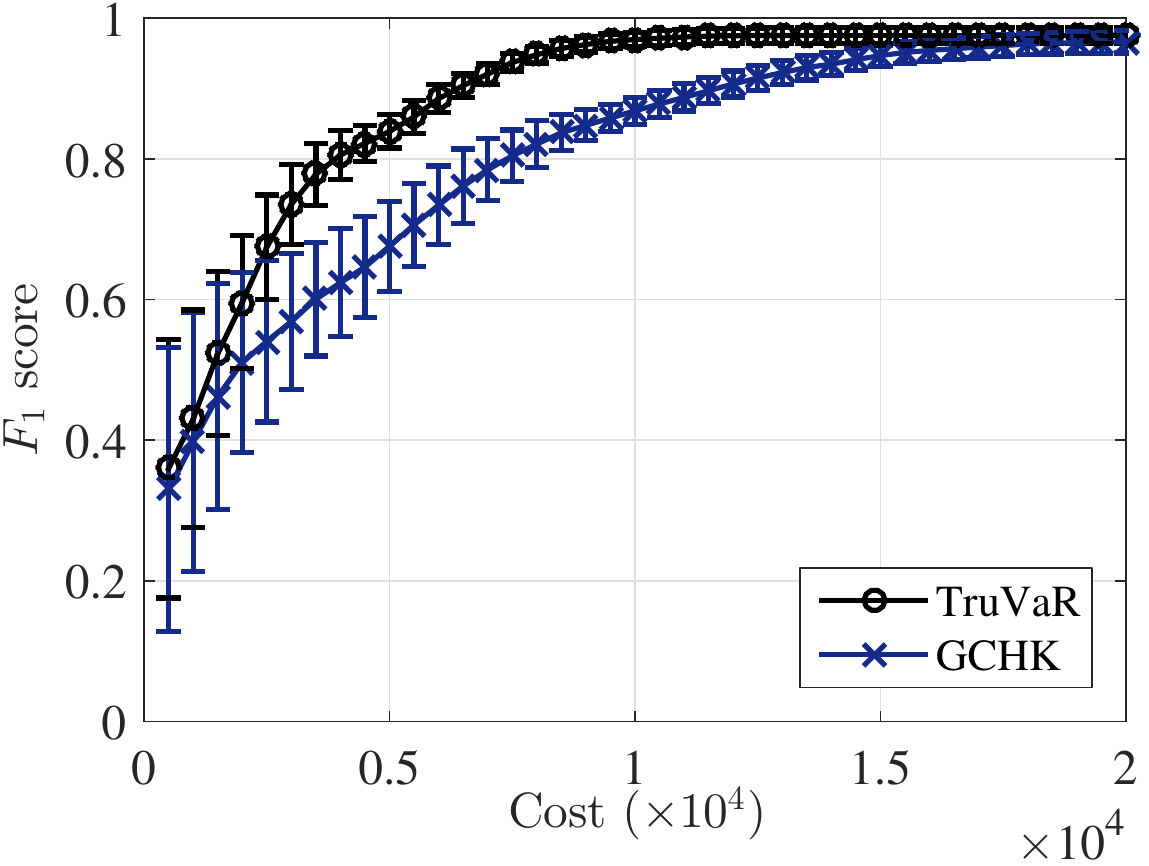}
		\caption{Lake data, varying cost} \label{fig:lake_cost}  
	\end{subfigure}
	\setcounter{subfigure}{2}
	\begin{subfigure}[b]{0.32\textwidth}
		\includegraphics[width=\textwidth]{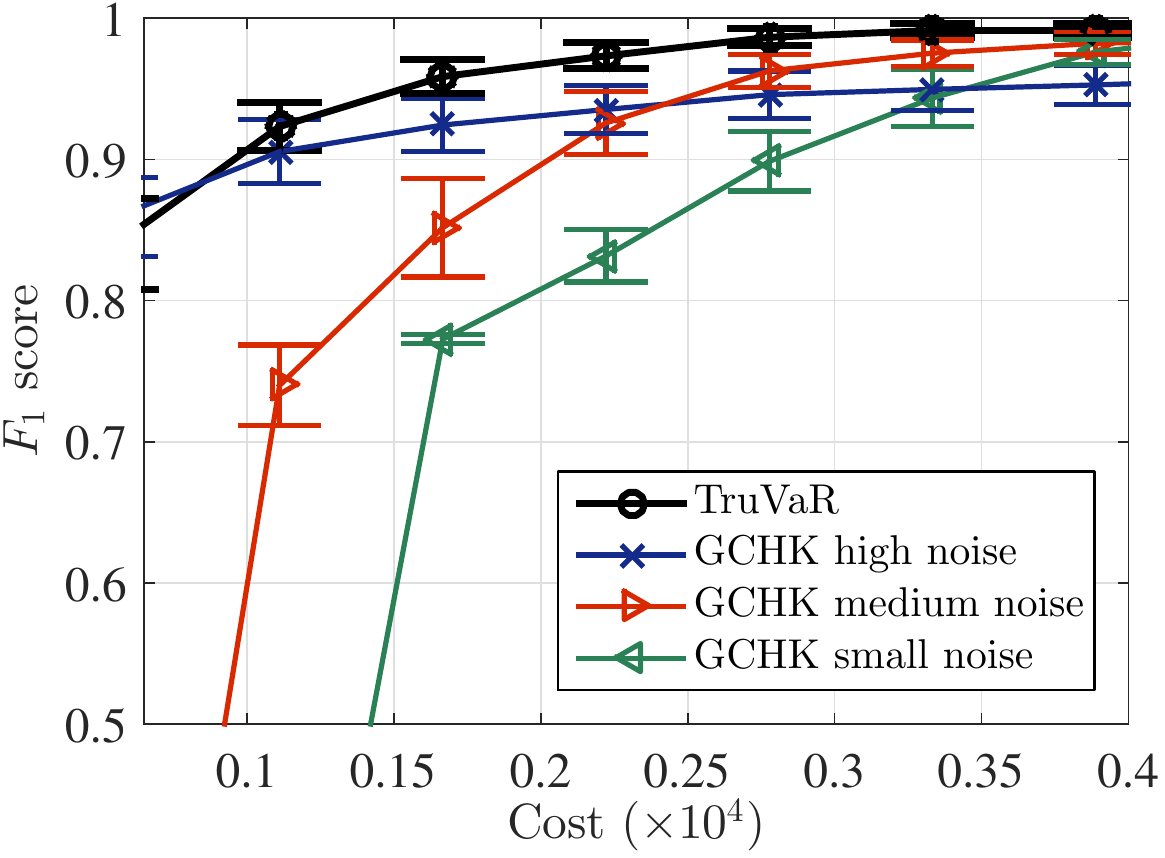}  
		\caption{Synthetic data, varying noise} \label{fig:choose_noise}
	\end{subfigure}
	\setcounter{subfigure}{3}
	\begin{subfigure}[b]{0.32\textwidth}
		\includegraphics[width=\textwidth]{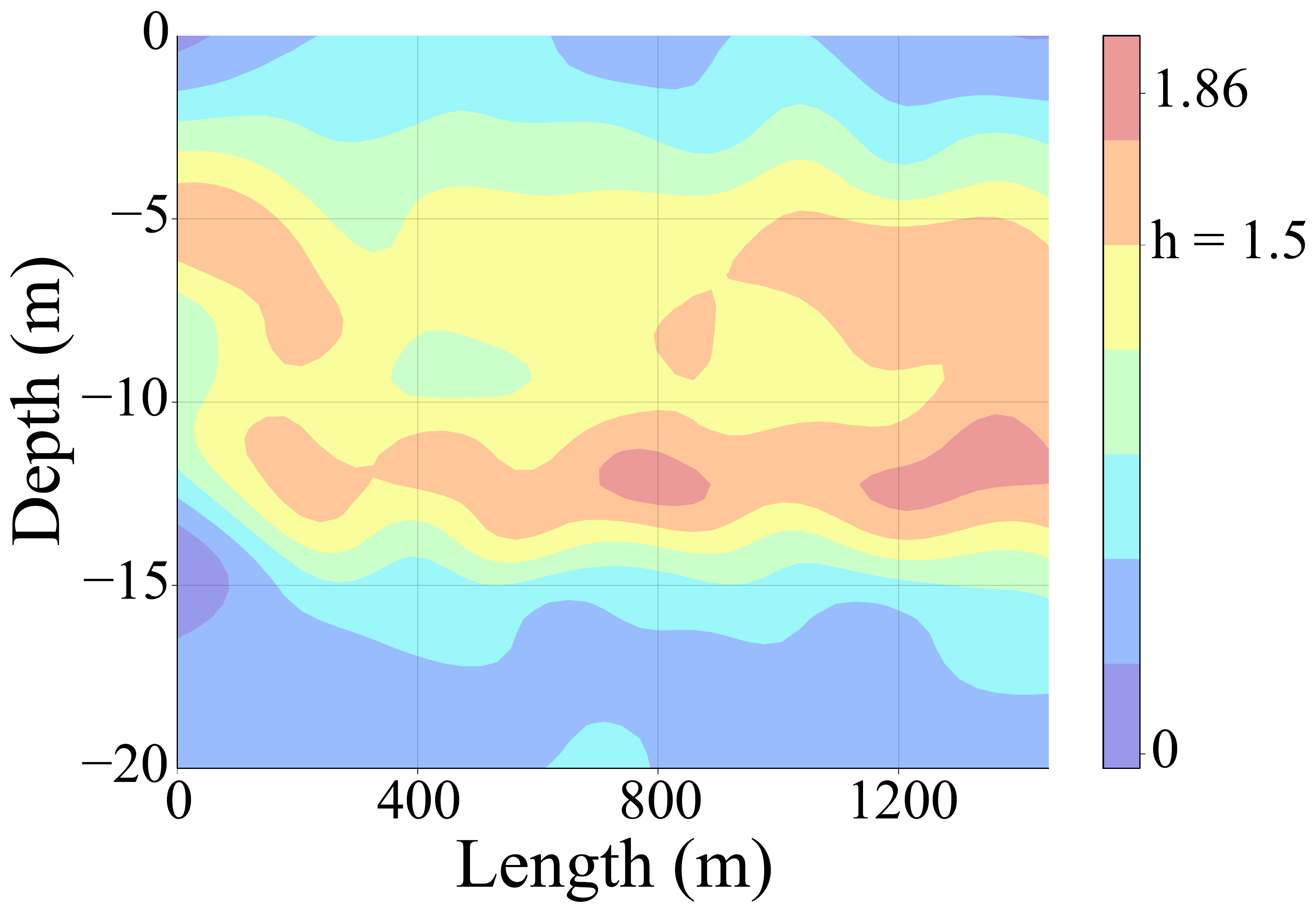}
		\caption{Inferred concentration function} 
		\label{fig:lake_f} 
	\end{subfigure}
	\setcounter{subfigure}{4}
	\begin{subfigure}[b]{0.32\textwidth}
		\includegraphics[width=\textwidth]{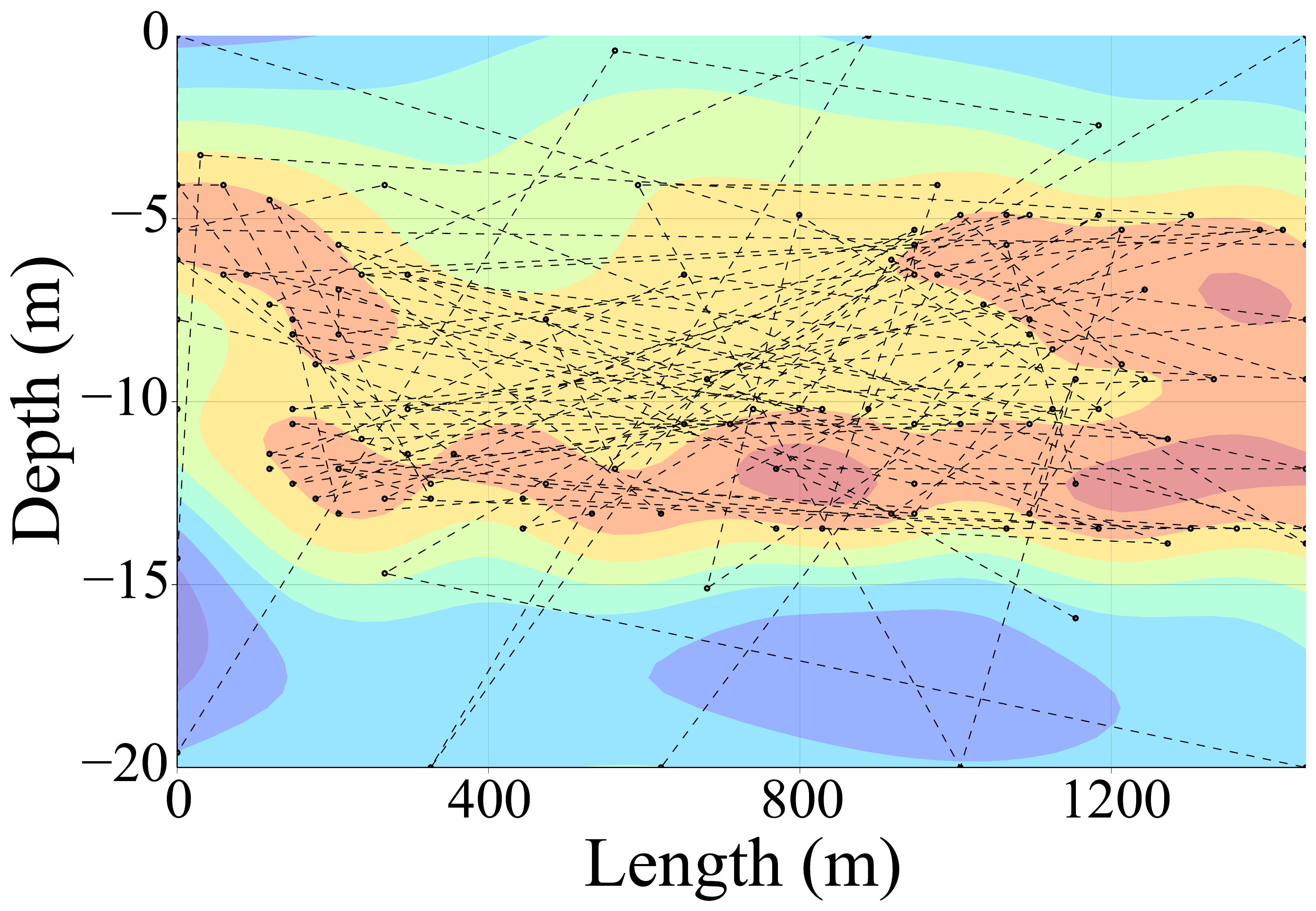}
		\caption{Points chosen by GCHK}  
		\label{fig:lse_run}	 
	\end{subfigure} 
	\setcounter{subfigure}{5}
	\begin{subfigure}[b]{0.32\textwidth}
		\includegraphics[width=\textwidth]{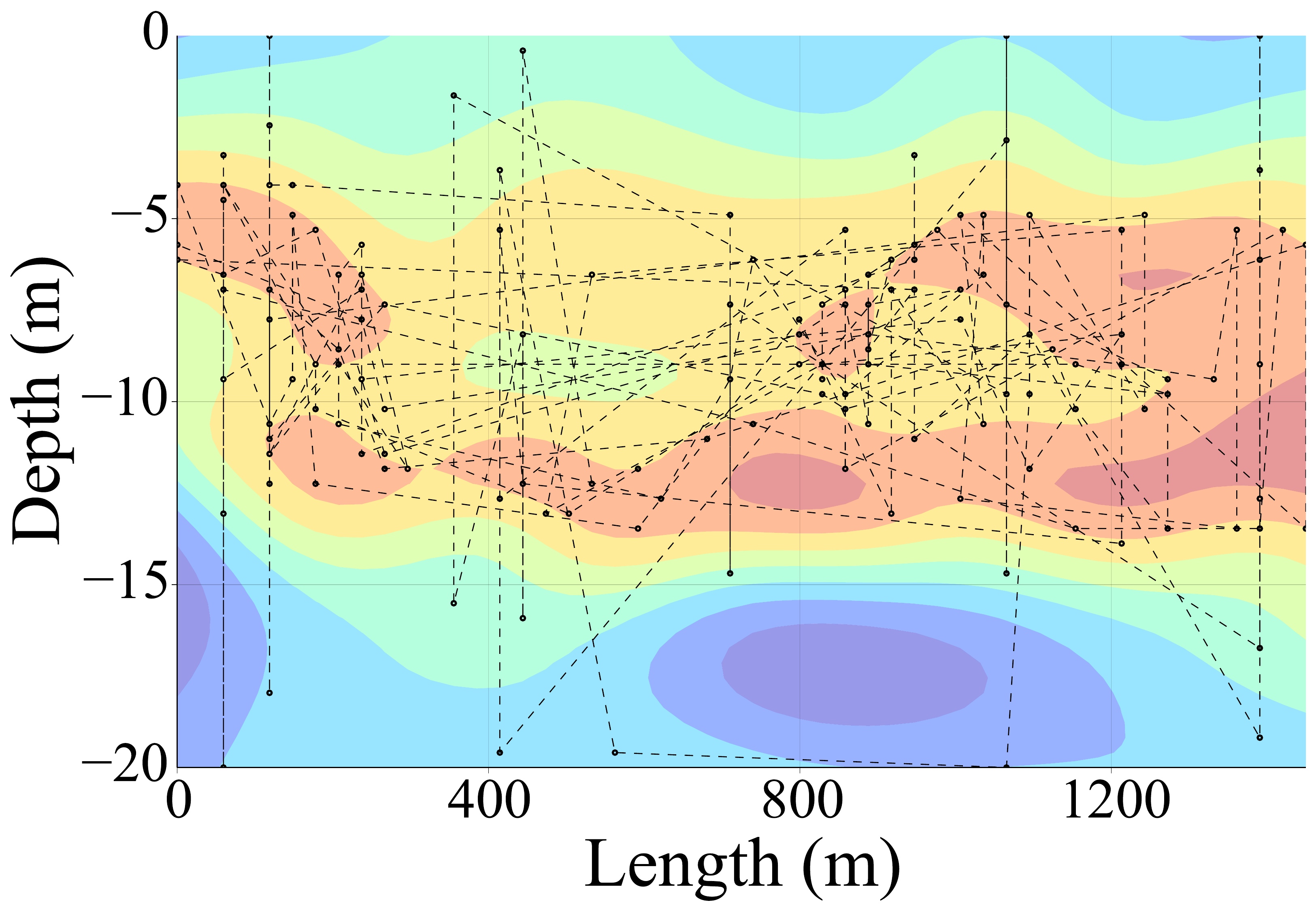}
		\caption{Points chosen by \ALGNAME} 
		\label{fig:truvar_run} 
	\end{subfigure}
	\caption{Experimental results for level-set estimation. \label{fig:main}} \vspace*{-2ex}
\end{figure}

{\bf Lake data (unit cost):} We begin with a data set from the domain of environmental monitoring of inland waters, consisting of 2024 in situ measurements of chlorophyll concentration within a vertical transect plane, collected by an autonomous surface vessel in Lake Z\"{u}rich~\cite{Htz2012}. As in~\cite{Got13}, our goal is to detect regions of high concentration. We evaluate each algorithm on a $50 \times 50$ grid of points, with the corresponding values coming from the GP posterior that was derived using the original data (see Figure~\ref{fig:lake_f}). We use the Mat\'{e}rn-5/2 ARD kernel, setting its hyperparameters by maximizing the likelihood on the second (smaller) available dataset.  The level-set threshold $h$ is set to $1.5$.  

In Figure~\ref{fig:lake_time}, we show the performance of the algorithms averaged over $100$ different runs; here the randomness is only with respect to the starting point, as we are in the noiseless setting.  We observe that in this unit-cost case, \ALGNAME~performs similarly to GCHK and STR.  All three methods outperform VAR, which is good for global exploration but less suited to level-set estimation.

{\bf Lake data (varying cost):} Next, we modify the above setting by introducing pointwise costs that are a function of the previous sampled point $x'$, namely, $c_{x'}(x) = 0.25|x_1 - x'_1| + 4(|x_2| + 1)$, where $x_1$ is the vessel position and $x_2$ is the depth.  Although we did not permit such a dependence on $x'$ in our original setup, the algorithm itself remains unchanged.  Our choice of cost penalizes the distance traveled $|x_1 - x'_1|$, as well as the depth of the measurement $|x_2|$.  Since incorporating costs into existing algorithms is non-trivial,  we only compare against the original version of GCHK that ignores costs.

In Figure~\ref{fig:lake_cost}, we see that TruVaR significantly outperforms GCHK, achieving a higher $F_1$ score for a significantly smaller cost.  The intuition behind this can be seen in Figures~\ref{fig:lse_run} and~\ref{fig:truvar_run}, where we show the points sampled by TruVaR and GCHK in one experiment run, connecting all pairs of consecutive points.  GCHK is designed to pick few points, but since it ignores costs, the distance traveled is large.  In contrast, by incorporating costs, \ALGNAME~tends to travel small distances, often even staying in the same $x_1$ location to take measurements at multiple depths $x_2$.


{\bf Synthetic data with multiple noise levels:} In this experiment, we demonstrate Corollary~\ref{thm:choose_noise} by considering the setting in which the algorithm can choose the sampling noise variance and incur the associated cost.  We use a synthetic function sampled from a GP on a $50 \times 50$ grid with an isotropic squared exponential kernel having length scale $l = 0.1$ and unit variance, and set $h=2.25$. We use three different noise levels, $\sigma^2 \in \{10^{-6}, 10^{-3}, 0.05\}$, with corresponding costs $\{15, 10, 2\}$.

We run GCHK separately for each of the three noise levels, while running \ALGNAME~as normal and allowing it to mix between the noise levels.  The resulting $F_1$-scores are shown in Figure \ref{fig:choose_noise}.  The best-performing version of GCHK changes throughout the time horizon, while \ALGNAME~is consistently better than all three.  A discussion on how \ALGNAME~mixes between the noise levels can be found in the supplementary material.

\begin{figure}
	\centering
	\setcounter{subfigure}{0}
	\begin{subfigure}[b]{0.325\textwidth}
		\includegraphics[width=\textwidth]{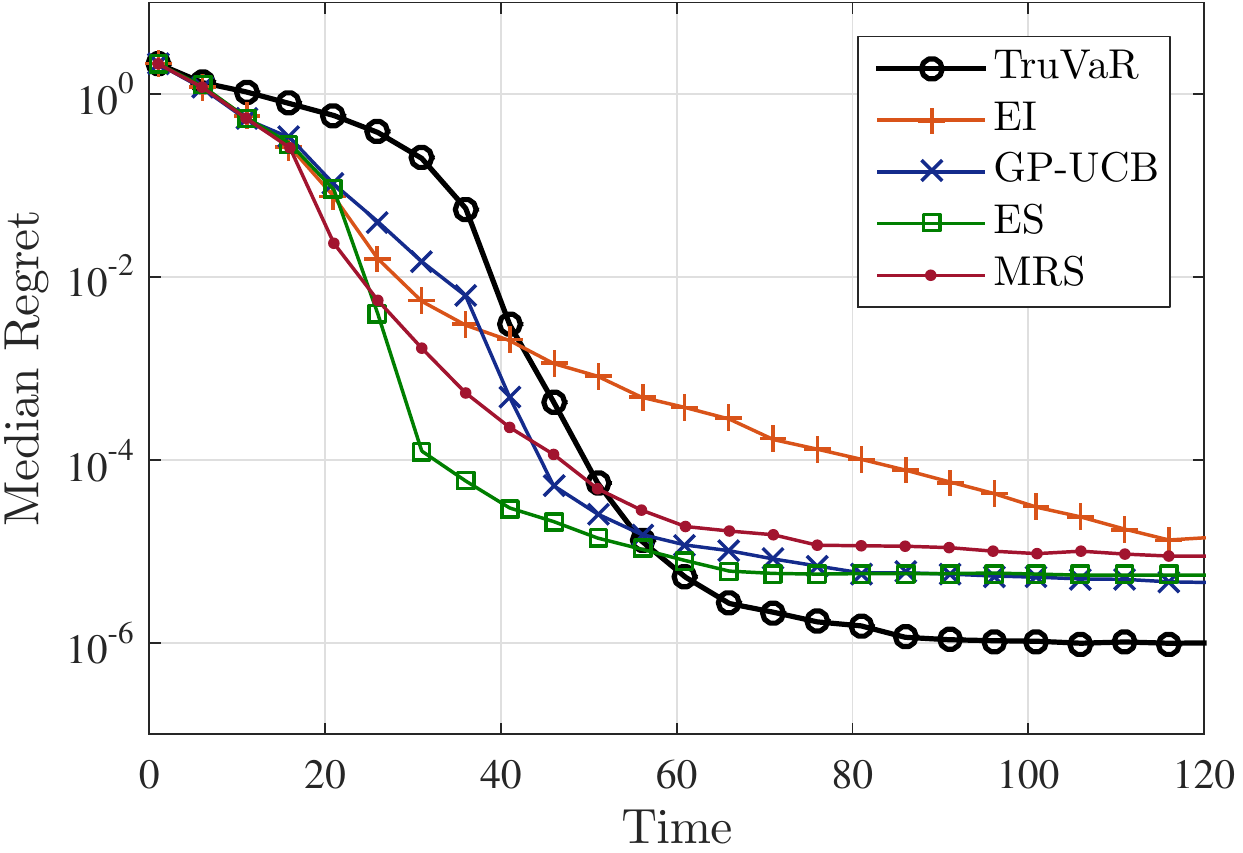}
		\caption{Synthetic, median}
		\label{fig:median_bo}
	\end{subfigure}
	\setcounter{subfigure}{1}
	\begin{subfigure}[b]{0.325\textwidth}
		\includegraphics[width=\textwidth]{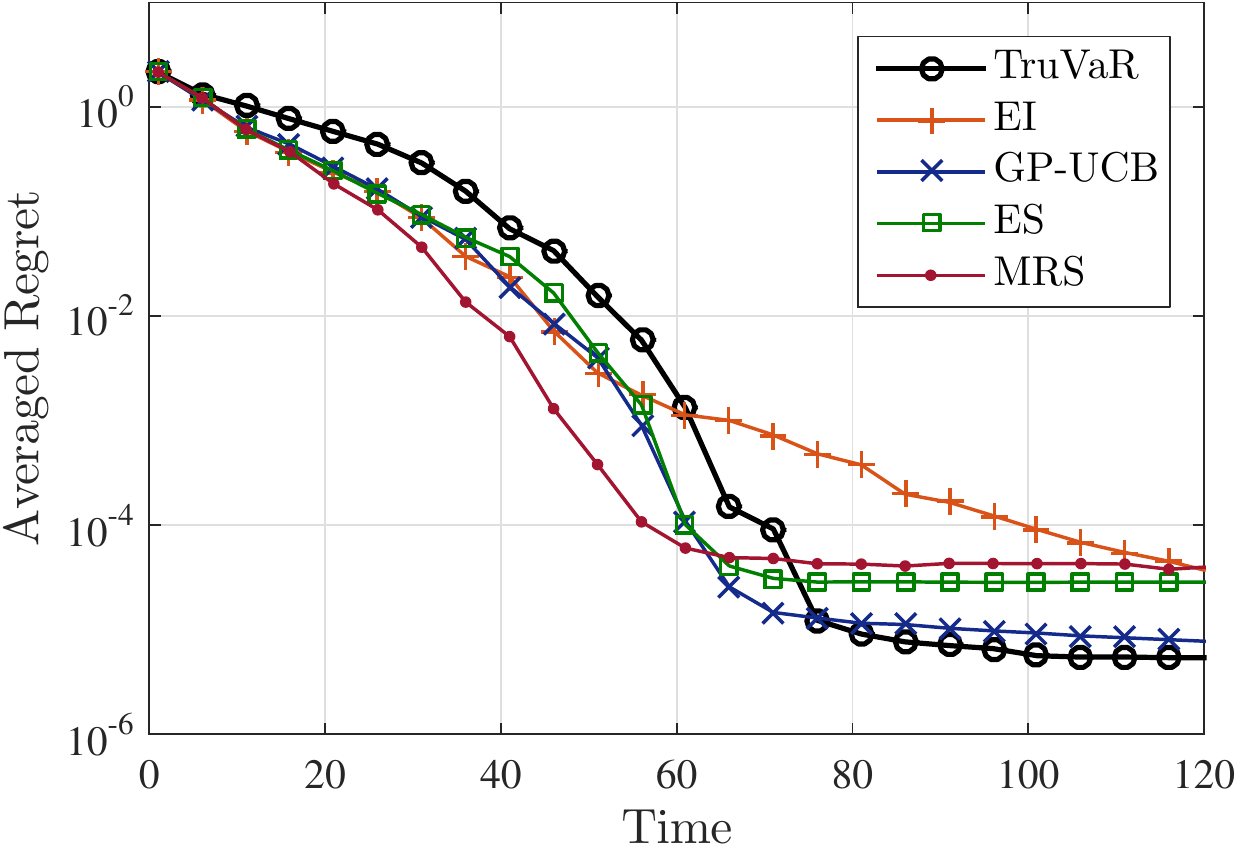}
		\caption{Synthetic,\,outlier-adjusted mean} 
		\label{fig:mean_bo}
	\end{subfigure}
	\setcounter{subfigure}{2}
	\begin{subfigure}[b]{0.325\textwidth}
		\includegraphics[width=\textwidth]{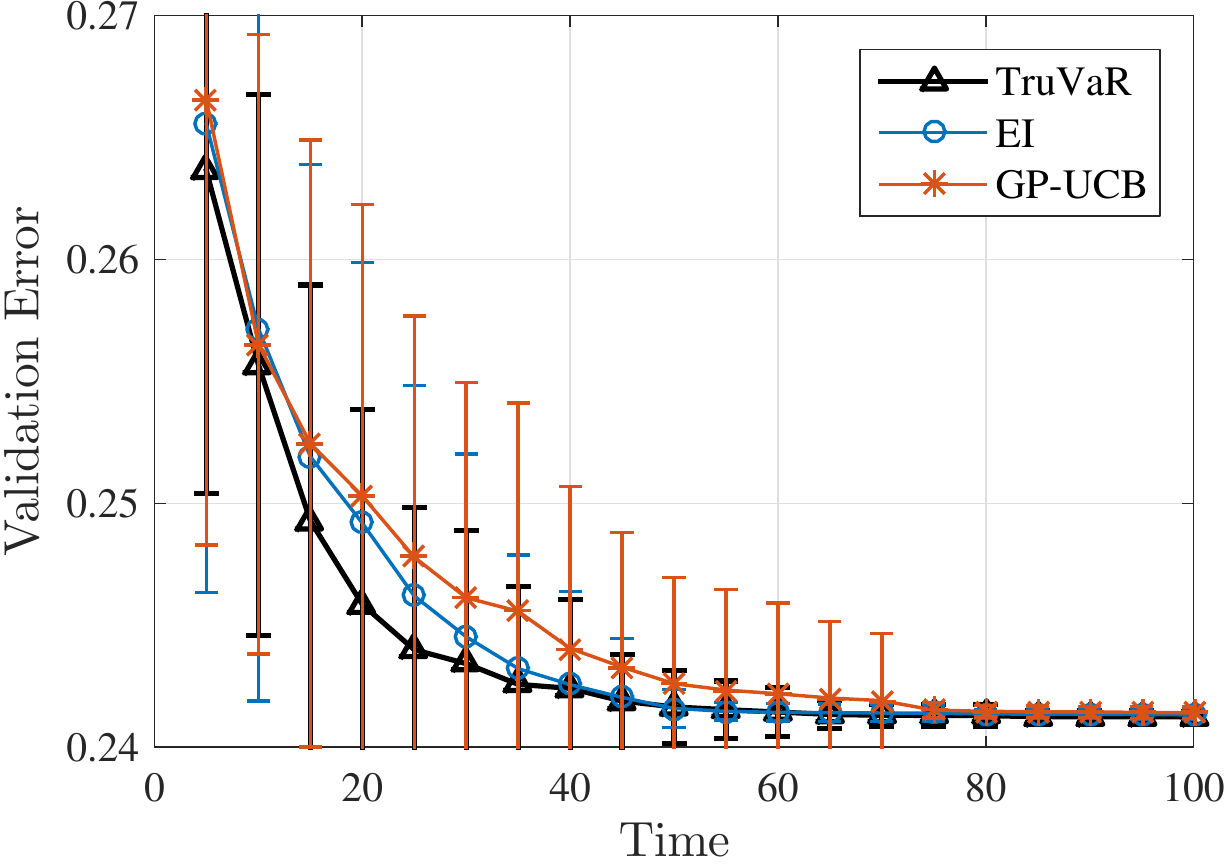}
		\caption{SVM data} 
		\label{fig:svm}  
	\end{subfigure}
	\caption{Experimental results for Bayesian optimization. \label{fig:bo}}
	\vspace*{-2ex}
\end{figure} 

\textbf{Bayesian optimization.} We now provide the results of two experiments for the BO setting.

 {\bf Synthetic data:} We first conduct a similar experiment as that in \cite{Hen12, Met16}, generating $200$ different test functions defined on $[0,1]^2$.  To generate a single test function, $200$ points are chosen uniformly at random from $[0,1]^2$, their function values are generated from a GP using an isotropic squared exponential kernel with length scale $l=0.1$ and unit variance, and the resulting posterior mean forms the function on the whole domain $[0,1]^2$.  We subsequently assume that samples of this function are corrupted by Gaussian noise with $\sigma^2=10^{-6}$.  The extension of \ALGNAME~to continuous domains is straightforward, and is explained in the supplementary material.   For all algorithms considered, we evaluate the performance according to the regret of a single reported point, namely, the one having the highest posterior mean.  
 

We compare the performance of \ALGNAME~against expected improvement (EI), GP-upper confidence bound (GP-UCB), entropy search (ES) and minimum regret search (MRS), whose acquisition functions are outlined in the supplementary material. We use publicly available code for ES and MRS \cite{ES_Code}. The exploration parameter $\beta_t$ in GP-UCB is set according to the recommendation in \cite{Sri12} of dividing the theoretical value by five, and the parameters for ES and MRS are set according to the recommendations given in \cite[Section 5.1]{Met16}. 

Figure~\ref{fig:median_bo} plots the median of the regret, and Figure \ref{fig:mean_bo} plots the mean after removing outliers (i.e., the best and worst 5\% of the runs).  In the earlier rounds, ES and MRS provide the best performance, while \ALGNAME~improves slowly due to exploration.  However, the regret of \ALGNAME~subsequently drops rapidly, giving the best performance in the later rounds after ``zooming in'' towards the maximum.  GP-UCB generally performs well with the aggressive choice of $\beta_t$, despite previous works' experiments revealing it to perform poorly with the theoretical value.


{\bf Hyperparameter tuning data:} In this experiment, we use the \textit{SVM on grid} dataset, previously used in~\cite{Sno12}. A $25 \times14 \times4$ grid of hyperparameter configurations resulting in $1400$ data points was pre-evaluated, forming the search space.  The goal is to find a configuration with small validation error.  We use a Mat\'{e}rn-5/2 ARD kernel, and re-learn its hyperparameters by maximizing the likelihood after sampling every $3$ points.  Since the hyperparameters are not fixed in advance, we replace $M_{t-1}$ by $D$ in \eqref{eq:Mt_opt} to avoid incorrectly ruling points out early on, allowing some removed points to be added again in later steps.  Once the estimated hyperparameters stop to vary significantly, the size of the set of potential maximizers decreases almost monotonically.  Since we consider the noiseless setting here, we measure performance using the simple regret, i.e., the best point found so far.

We again average over $100$ random starting points, and plot the resulting validation error in Figure~\ref{fig:svm}.  Even in this noiseless and unit-cost setting that EI and GP-UCB are suited to, we find that \ALGNAME~performs slightly better, giving a better validation error with smaller error bars.

\section{Conclusion}

We highlight the following aspects in which \ALGNAME~is versatile:
\begin{itemize}[leftmargin=3ex, topsep=0ex] \itemsep0em 
    \item \textbf{Unified optimization and level-set estimation:} These are typically treated separately, whereas \ALGNAME~and its theoretical guarantees are essentially identical in both cases
    \item \textbf{Actions with costs:} \ALGNAME~naturally favors cost-effective points, as this is directly incorporated into the acquisition function.  
    \item \textbf{Heteroscedastic noise:} \ALGNAME~chooses points that effectively shrink the variance of \emph{other} points, thus directly taking advantage of situations in which some points are noisier than others.  
    \item \textbf{Choosing the noise level:} We provided novel theoretical guarantees for the case that the algorithm can choose both a point and a noise level, \emph{cf.}, Corollary \ref{thm:choose_noise}.  
\end{itemize}
Hence, \ALGNAME~directly handles several important aspects that are non-trivial to incorporate into myopic algorithms.  Moreover, compared to other BO algorithms that perform a lookahead (e.g., ES and MRS), \ALGNAME~avoids the computationally expensive task of averaging over the posterior and/or measurements, and comes with rigorous theoretical guarantees. 

\textbf{Acknowledgment:} This work was supported in part by the European Commission under Grant ERC Future Proof, SNF Sinergia project CRSII2-147633, SNF 200021-146750, and EPFL Fellows Horizon2020 grant 665667.


\newpage
\bibliographystyle{IEEEtran}
\bibliography{refs}

\appendix

\newpage
{\huge \bf Supplementary Material}

{\large \bf Truncated Variance Reduction: A Unified Approach to Bayesian Optimization and Level-Set Estimation} {\large (Ilija Bogunovic, Jonathan Scarlett, Andreas Krause, and Volkan Cevher, NIPS 2016) }

\section{Variations of the \ALGNAME~Algorithm} \label{sec:variations}

Our algorithm \ALGNAME~can naturally be adapted to suit various settings, including the following:
\begin{itemize}[leftmargin=3ex]
    \item \textbf{Non-monotonic $M_t$:} We have defined our sets $M_t$ to become smaller on every time step.  However, if $\beta_{(i)}$ is chosen aggressively, it may be preferable to replace $M_{t-1}$ by $D$ in \eqref{eq:Mt_opt}--\eqref{eq:Mt_lvl}, in which case some removed points may be added back in depending on how the posterior mean changes between steps.  We take this approach in the real-world BO example of Section \ref{sec:numerical} in which the kernel hyperparameters are learned online, so as to avoid incorrectly ruling out points early on due to mismatched hyperparameters.
    \item \textbf{Avoiding computing the acquisition function everywhere:} We found that instead of computing the acquisition function at every point in $D$, limiting the selection to points in $M_{t-1}$ has minimal effect on the performance.  To reduce the computation even further, one could adopt a strategy such as that proposed in \cite{Swe14}: Take some relatively small number of points having the top GP-UCB or EI score, and then choose the point \emph{in that restricted subset} having the highest score according to \eqref{eq:acq}.  In fact, the numerical results in Figure \ref{fig:bo} suggest that this may not only reduce the computation, but also improve the performance in the very early rounds by making the algorithm \emph{initially} behave more like GP-UCB or EI.
    \item \textbf{Pure variance reduction:} Setting $\eta_{(1)} = 0$ yields a \emph{pure variance reduction} algorithm, which minimizes the total variance within $M_t$ via a one-step lookahead.  While our theory does not apply in this case, we found this choice to also work well in practice.
    \item \textbf{Implicit threshold for level-set estimation:} While we have focused on a threshold $h$ for level-set estimation that is fixed in advance, one can easily incorporate the ideas of \cite{Got13} to allow for an \emph{implicit} threshold which is equal to some constant multiple of the function's maximum, which is random and unknown in advance.
    \item \textbf{Anticipating changes in $M_t$:} The acquisition function \eqref{eq:acq} computes the truncated variance reduction resulting from a one-step lookahead, but still sums over the previous set $M_{t-1}$.  In order to make it more preferable to choose points that shrink $M_t$ faster, it may be preferable to instead sum over $M_{t-1|x}$, defined to be the updated set upon adding $x$.  The problem here is that such an update depend on the next posterior mean, whose update requires sampling $f$.  One solution is to average over the measurement as in \cite{Hen12}; alternatively, a simpler approach is to replace the observed value with its mean when doing this one-step lookahead, and then using the true observed function sample only when $x_t$ is actually chosen.
   \item \textbf{Continuous domains:} Our algorithm also extends to compact domains such as $[0,1]^d$.  The main challenge is that the summations in \eqref{eq:acq} become integrals that need to be approximated numerically.  The simplest way of  doing this is to approximate the integrals by summations over a finite number of \emph{representer points}, e.g., a grid of values that cover the domain sufficiently densely.  The theoretical analysis of this modified algorithm is left for future work.
    \item \textbf{Batch setting:} As we show in the proof of Theorem \ref{thm:general}, our algorithm can be interpreted as performing the first step of a greedy submodular covering problem at each time step.  This leads to a very natural extension to the batch setting, in which multiple points (say, $k$ of them) are chosen at each time step: Simply perform the first $k$ steps of the greedy covering algorithm during each batch.
\end{itemize}

\section{Further Details of Numerical Experiments}

\textbf{Other algorithms considered:} We outline the algorithms that \ALGNAME~is compared against; full details can be found in the cited papers.  For level-set estimation, we have the following:
\begin{itemize}
    \item The GCHK algorithm \cite{Got13} evaluates, at each iteration, the point that is not yet classified with the largest ambiguity: $x_t = \argmax_{x \in M_{t-1}} \min\{u_t(x) - h, h - \ell_t(x)\}$, where $u_t$ and $\ell_t$ are defined as in \eqref{eq:confidence_bounds} with a parameter $\beta_t$ replacing $\beta_{(i)}$.  Here, similarly to our algorithm, $M_{t-1}$ is the set of points that have not yet been classified as having a value above or below the threshold $h$.
    \item The straddle (STR) heuristic~\cite{Bry08} chooses $x_t = \argmax_{x \in D} 1.96  \sigma_{t - 1}(x) - |\mu_{t-1}(x) - h|$, favoring high-uncertainty points that are expected to have function values closer to $h$.
    \item The maximum variance rule (VAR) \cite{Got13} simply chooses $x_t = \argmax_{x \in D} \sigma_{t-1}(x)$.
\end{itemize}
For Bayesian optimization, we have the following:
\begin{itemize}
    \item The expected improvement (EI) algorithm \cite{Sha16} chooses $x_t = \argmax_{x \in D} \myexpect_t[(f(x) - \xi_t)\mathds{1}\{ f(x) > \xi_t \}]$, where $\myexpect_t[\cdot]$ denotes averaging with respect to the posterior distribution at time $t$, and $\xi_t$ is the best observed value so far.  Since the posterior is Gaussian, the expectation can easily be expressed in closed form.
    \item The Gaussian Process Upper Confidence Bound (GP-UCB) algorithm \cite{Sri12} chooses the points with the highest upper confidence bounds, $x_t = \argmax_{x \in D} \mu_{t-1}(x) + \beta_t \sigma_{t-1}(x)$, where $\beta_t$ is a parameter controlling the level of exploration performed.
    \item The Entropy Search (ES) algorithm can be interpreted as approximating the rule $x_t = \argmin_{x \in D} h(f_{x^*|t,x})$, where $h(f) = \int_{D} f(x)\log\frac{1}{f(x)}dx$ denotes the differential entropy, and $f_{x^*|t,x}$ denotes the density function of the optimal action $x^*$ given the observations up to time $t$ along with the additional observation $x$.  Intuitively, this rule seeks to minimize the uncertainty of $x^*$.  Since its exact evaluation is intractable, it is approximated using Monte Carlo techniques to average with respect to the posterior distribution and the measurements.
    \item The Minimum Regret Search (MRS) algorithm \cite{Met16} also resembles ES, but works with the expected regret instead of the differential entropy.  Once again, Monte Carlo techniques are used to average with respect to the posterior distribution and the measurements.
\end{itemize}

\begin{figure}
	\centering
	\setcounter{subfigure}{0}
	\begin{subfigure}[b]{0.32\textwidth}
		\includegraphics[width=\textwidth]{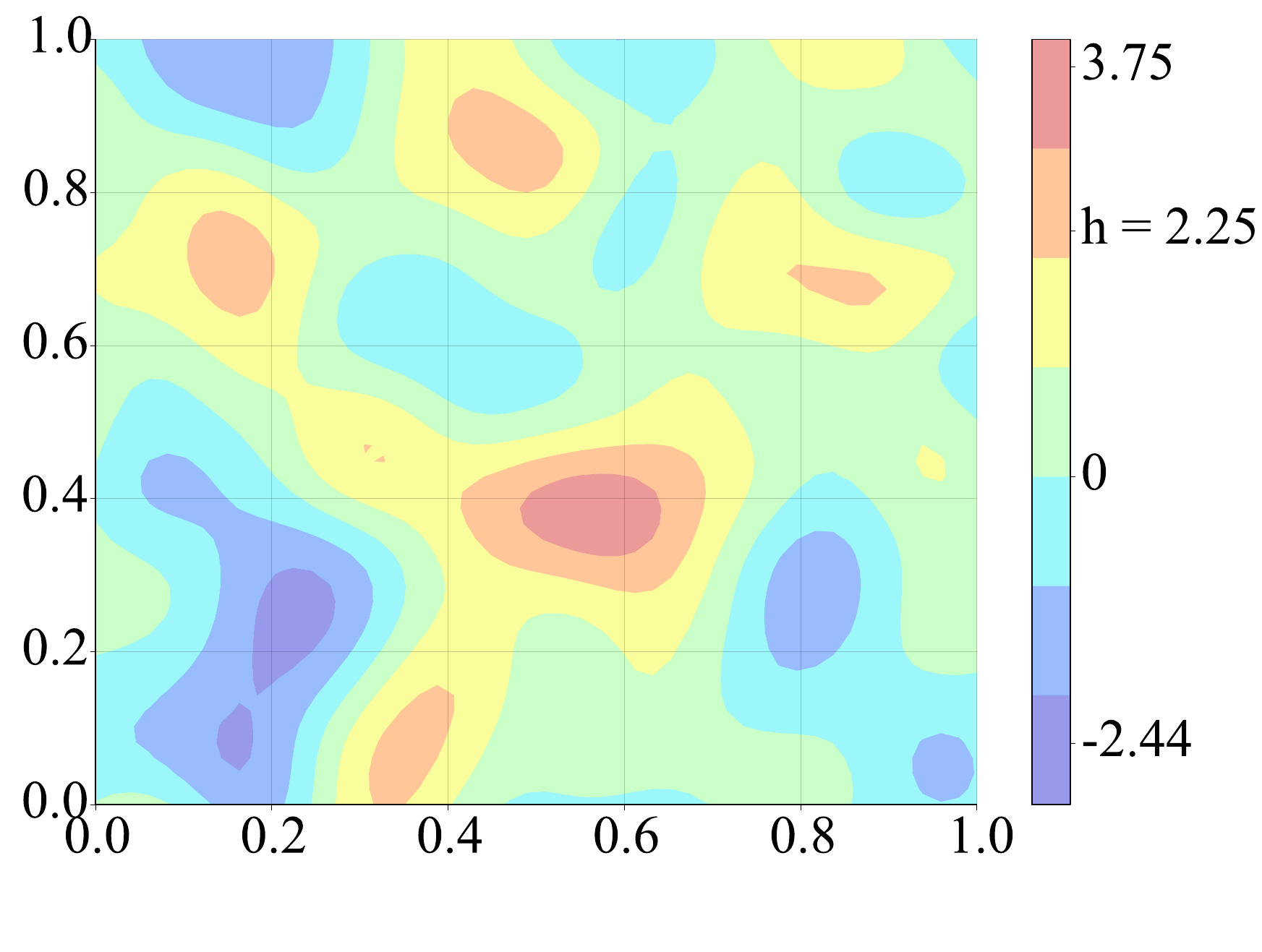}
		\caption{}
		\label{fig:lse_synth}
	\end{subfigure}
	\setcounter{subfigure}{1}
	\begin{subfigure}[b]{0.355\textwidth}
		\includegraphics[width=\textwidth]{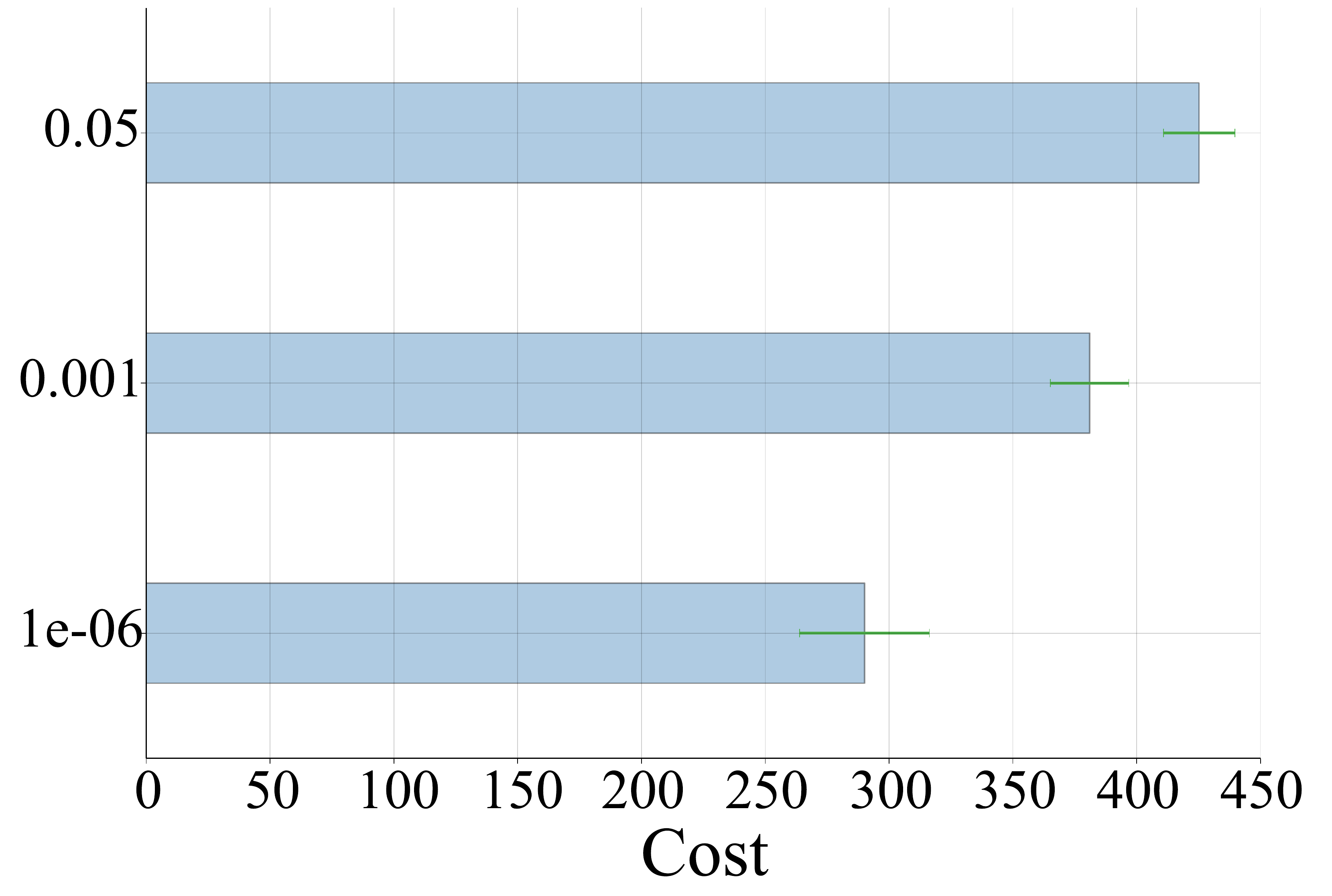} 
		\caption{} 
		\label{fig:noise_hist}
	\end{subfigure}
	\caption{(a) Function used in synthetic level-set estimation experiments; (b) The total cost used by \ALGNAME~for each of the three noise levels.}
	\vspace*{-2ex}
\end{figure} 

\textbf{Efficiently computing the acquisition function:}
To compute the value of the acquisition function~\eqref{eq:acq} for different $x \in D$, we need to compute $\sigma_{t-1|x}^2(M_{t-1}) \in \mathbb{R}^{|M_{t-1}|}$, i.e., the posterior variance of points in $M_{t-1}$ upon observing $x$ along with $x_1,\cdots,x_{t-1}$.
Instead of computing it directly, it is more efficient to recursively compute $\sigma_{t-1|x}^2(M_{t-1}) = \sigma_{t-1}^2(M_{t-1}) - \Delta_{t-1|x}(M_{t-1})$. The difference term, $\Delta_{t-1|x}(M_{t-1})$, can be computed as \cite{Hen12}:
\begin{equation}
	\Delta_{t-1|x}(M_{t-1}) = \diag \big(\text{Cov}_{t-1}(M_{t-1}, x) (\sigma^2 + \sigma^2_{t-1}(x))^{-1} \text{Cov}_{t-1}(M_{t-1},x)^T\big),
\end{equation}	
 where 
\begin{align}
	\sigma_{t-1}^2(x) &= k(x,x) - \vk_{t-1}(x)^T \big(\vK_{t-1} + \vSigma_{t-1} \big)^{-1} \vk_{t-1}(x)\label{eq:var_eq} \\ 
	\text{Cov}_{t-1}(M_{t-1}, x)   &= \vk(M_{t-1}, x) - \vk_{t-1}(M_{t-1})^T \big(\vK_{t-1} + \vSigma_{t-1} \big)^{-1} \vk_{t-1}(x), \label{eq:cov_eq}
\end{align}
and where $\vk(M_{t-1}, x) = \big[k(\xbar, x)\big]_{\xbar \in |M_{t-1}|} \in \mathbb{R}^{|M_{t-1}|}$, and $\vk_{t-1}(M_{t-1}) = \big[k(\xbar, x)\big]_{\xbar\in|M_{t-1}|, x\in\{1,\dotsc,t-1\}} \in \mathbb{R}^{|M_{t-1}| \times (t-1) }$. 
When the Cholesky decomposition of $\vK_{t-1} + \vSigma_{t-1}$ is known, $\big(\vK_{t-1} + \vSigma_{t-1} \big)^{-1} \vk_{t-1}(x)$ can be computed in time $O(t^2)$. 

\textbf{Details on LSE experiment with multiple noise levels:} 
Figure \ref{fig:lse_synth} plots the randomly-generated function that was used in this experiment.  Figure \ref{fig:noise_hist} plots the average cost spent by \ALGNAME~on each noise level by the end of the experiment, again averaged over 100 trials.  We see that the cost is roughly equally distributed across the three levels.  To be more specific, we observed that \ALGNAME~initially chooses high noise levels in order to cheaply explore, and throughout the course of the experiments, it gradually switches to lower noise levels in order to accurately determine the function values around the maximum.  This is consistent with the behavior of the three version of GCHK, with $\sigma^2 = 0.05$ performing well in the early stages, but $\sigma^2 = 10^{-6}$ being preferable in the later stages.

\textbf{Extension of \ALGNAME~for synthetic BO experiment} We used the extension of \ALGNAME~to continuous domains outlined in Appendix \ref{sec:variations}, approximating the integrals over $M_t$ by summations that are restricted to points on a uniformly-spaced $50 \times 50$ grid covering $[0,1]^2$.  We optimized our acquisition function using DIRECT \cite{Jon93}.


\section{Proof of General Result (Theorem \ref{thm:general})} \label{sec:greedy_proof}

We begin with the following lemma from \cite{Sri12}.

\begin{lemma} \emph{\cite{Sri12}} \label{lem:conf_bounds}
    For each $t$, define $\beta_{t} = 2\log\frac{|D| t^2 \pi^2 }{ 6\delta }$.  With probability at least $1 - \delta$, we have for all $x$ and $t$ that $|f(x) - \mu_t(x)| \le \beta_{t}^{1/2}\sigma_t(x)$.
\end{lemma}

We conclude that in order for $\mu_t(\cdot) \pm \beta_{(i)}^{1/2} \sigma_t(\cdot)$ to provide valid confidence bounds, it suffices to ensure that $\beta_{(i)} \ge \beta_t$ for all $t$ in epoch $i$.  From \eqref{eq:beta_i}, we see that this is true provided that the time taken to reach the end of the $i$-th epoch is at most $\frac{1}{\cmin}\sum_{i' \le i} C_{(i')}$.  Since $\cmin$ is the minimum pointwise cost, this holds provided that the cost incurred in epoch $i$ is at most $C_{(i)}$.  The bulk of the proof is devoted to showing that this is the case.

\begin{figure}
    \begin{centering}
        \includegraphics[width=0.4\columnwidth]{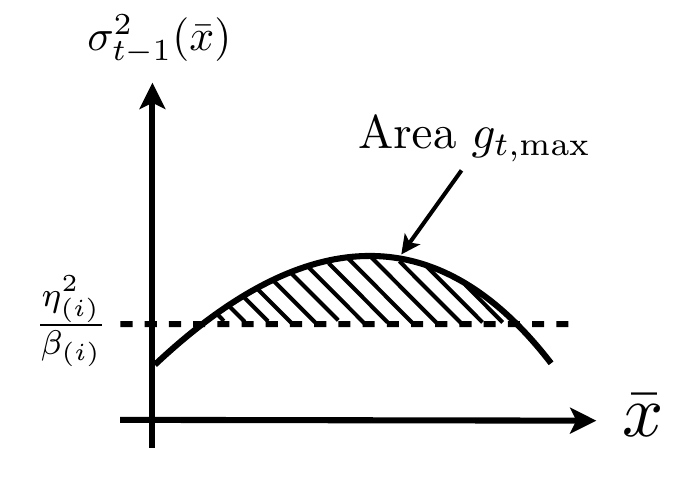}
        \par
    \end{centering}
    
    \caption{Illustration of the excess variance $\gtmax$. \label{fig:gtmax}}
\end{figure}

We connect \ALGNAME~with the following budgeted submodular covering problem:\footnote{Recall that $S$ may contain duplicates, and these are counted multiple times accordingly in the definitions of both $c(S)$ and $g_t(S)$.  All of our equations can be cast in terms of standard sets (without duplicates) by expanding $D$ to $D \times \{1,\cdots,N\}$ for any integer $N$ that is larger than the maximum number of points that are chosen throughout the course of the algorithm. }
\begin{equation}
    \text{minimize}_S ~~ c(S) \quad \text{ subject to }~ g_t(S) = \gtmax, \label{eq:submod_opt}
\end{equation}

where
\begin{equation}
    g_t(S) = \sum_{\xbar\in M_{t-1}}\max\bigg\{ \sigma_{t-1}^2(\xbar), \frac{\eta^2_{(i)}}{\beta_{(i)}} \bigg\} - \sum_{\xbar\in M_{t-1}}\max\bigg\{ \sigma_{t-1|S}^2(\xbar), \frac{\eta^2_{(i)}}{\beta_{(i)}} \bigg\}, \label{eq:g_t}
\end{equation}
and where $\gtmax$ is the highest possible value of $g_t(S)$ over arbitrarily large $S$, i.e., it is the value obtained once all of the summands in the second summation in \eqref{eq:g_t} have saturated to $\frac{\eta^2_{(i)}}{\beta_{(i)}}$:
\begin{align}
    \gtmax 
        &= \sum_{\xbar\in M_{t-1}} \bigg(\max\bigg\{ \sigma_{t-1}^2(\xbar), \frac{\eta^2_{(i)}}{\beta_{(i)}} \bigg\} - \frac{\eta^2_{(i)}}{\beta_{(i)}}\bigg) \label{eq:gtmax} \\
        &= \sum_{\xbar \in M_{t-1}} \max\bigg\{0,\sigma_{t-1}^2(\xbar) - \frac{\eta^2_{(i)}}{\beta_{(i)}} \bigg\}. \label{eq:gtmax2}
\end{align}
We thus refer to $\gtmax$ as the \emph{excess variance}; see Figure \ref{fig:gtmax} for an illustration.  Note that each time instant $t$ corresponds to a different function $g_t(S)$, and we are considering sets $S$ of an arbitrary size even though our algorithm only chooses one point at each time instant.

By our assumption on the submodularity of the variance reduction function, and the fact that taking the minimum with a constant\footnote{The minimum becomes a maximum after negation.} preserves submodularity \cite{Kra12}, $g_t(S)$ is also submodular.  It is also easily seen to be monotonically increasing, and normalized in the sense that $g_t(\emptyset) = 0$. 

Our selection rule \eqref{eq:acq} at time $t$ can now be interpreted as the first step in a greedy algorithm for solving the budgeted submodular optimization problem \eqref{eq:submod_opt}; specifically, the greedy rule optimizes the objective value per unit cost.  To obtain performance guarantees, we use Lemma 2 of \cite{Kra05} specialized to $|S| = 1$, which reads as follows in our own notation:
\begin{equation}
    g_t(\{x_t\}) \ge \frac{c(x_t)}{c(S_t^*)} \gtmax, \label{eq:greedy_bound}
\end{equation}
where $S_t^*$ is an optimal solution to \eqref{eq:submod_opt}, and hence $g_t(S^*) = \gtmax$.  Here $x_t$ is the point chosen greedily by our algorithm.

We now consider the behavior of the excess variance $\gtmax$ in a single epoch, i.e., the duration of a single value of $i$ in the algorithm.  We claim that for $t$ and $t+1$ in the same epoch, we have
\begin{equation}
    \gtimax \le \gtmax - g_t(\{x_t\}). \label{eq:gt_ineq}
\end{equation}
To see this, we note from \eqref{eq:g_t}--\eqref{eq:gtmax} that this would hold with equality if we were to have $M_{t} = M_{t-1}$, since by definition we have $\sigma_{t}^2(\xbar) = \sigma_{t-1|\{x_t\}}^2(\xbar)$.  We therefore obtain \eqref{eq:gt_ineq} by recalling that $M_{t}$ is decreasing in $t$ with respect to inclusion, and noting from \eqref{eq:gtmax2} that any given $\gtmax$ can only decrease when $M_{t-1}$ is smaller.

Combining \eqref{eq:greedy_bound}--\eqref{eq:gt_ineq} gives
\begin{equation}
    \gtimax \le \bigg( 1 - \frac{c(x_t)}{c(S_t^*)} \bigg) \gtmax.  \label{eq:gt_init}
\end{equation}
We also note that $c(S_{t+1}^*) \le c(S_t^*)$, which follows since $\sigma_{t}^2(\cdot)$ is decreasing in $t$ and $M_{t}$ is shrinking in $t$, and therefore at time $t+1$ a smaller cost is required to ensure that all terms in the second summation of \eqref{eq:g_t} have saturated to $\frac{\eta^2_{(i)}}{\beta_{(i)}}$.  Hence, and applying \eqref{eq:gt_init} recursively, we obtain for $t$ and $t+\ell$ in the same epoch that
\begin{align}
    \frac{\gtsummax}{\gtmax} 
        &\le \prod_{t'=t+1}^{t+\ell} \bigg(1 - \frac{c(x_{t'})}{c(S_t^*)}\bigg) \\
        &\le \exp\bigg(- \frac{\sum_{t'=t+1}^{t+\ell}c(x_{t'})}{c(S_t^*)}\bigg),
\end{align}
where we have applied the inequality $1-\alpha \le e^{-\alpha}$.  Moreover, the total cost incurred by choosing these points is precisely $\sum_{t'=t+1}^{t+\ell}c(x_t)$.  Thus, letting $t_{(i)}$ be the first time index in the $i$-th epoch, we find that in order to remove all but a proportion $\gamma$ of the initial excess variance $\gtomax$, it suffices that the cost incurred is at least
\begin{equation}
    c(S_{t_{(i)}}^*)\log\frac{1}{\gamma}. \label{eq:gamma_cost}
\end{equation}

Next, we observe that since the posterior variance is upper bounded by one due to the assumption $k(x,x) = 1$, the initial excess variance $\gtomax$ is upper bounded by $\gtomax \le |M_{t_{(i)}-1}|$, the size of the set of potential maximizers at the \emph{start} of the epoch. It follows that if we set 
\begin{equation}
    \gamma = \frac{\deltabar^2\eta_{(i)}^2 }{ \beta_{(i)} |M_{t_{(i)} - 1}|}, \label{eq:gamma_choice}
\end{equation}
then removing all but a proportion $\gamma$ of $\gtomax$ also implies removing all but $\deltabar^2\frac{\eta_{(i)}^2}{\beta_{(i)}}$ of it.  In other words, if at time $t$ we have incurred a cost in epoch $i$ satisfying \eqref{eq:gamma_cost} with $\gamma$ as in \eqref{eq:gamma_choice}, then we must have $\gtmax \le \deltabar^2\frac{\eta_{(i)}^2}{\beta_{(i)}}$.

Removing all of the excess variance would imply $\eta_{(i)}$-confidence at all points in $M_t$.  In the worst case, the remaining excess variance $\deltabar^2\frac{ \eta_{(i)}^2}{\beta_{(i)}}$ is concentrated entirely on a single point, in which case its confidence is upper bounded by $\sqrt{1+\deltabar^2}\eta_{(i)}$, which is further upper bounded by $(1+\deltabar)\eta_{(i)}$ due to the identity $\sqrt{1+\alpha^2} \le 1+\alpha$.

Combining these observations, we conclude that in the $i$-th epoch, upon incurring a cost of at least
\begin{equation}
    c(S_{t_{(i)}}^*)\log\frac{|M_{t_{(i)} - 1}| \beta_{(i)}}{\deltabar^2 \eta_{(i)}^2}, \label{eq:Ci_LB0}
\end{equation}
we are guaranteed to have $(1+\deltabar)\eta_{(i)}$-confidence for all points in $M_t$.  Having such confidence is precisely the condition used in the algorithm to move onto the next epoch, and we conclude that the epoch must end by (or sooner than) the time that \eqref{eq:Ci_LB0} holds.  

In accordance with the discussion following Lemma \ref{lem:conf_bounds}, we need to show that when the high-probability event in that lemma holds true, \eqref{eq:Ci_LB0} is upper bounded by the right-hand side of \eqref{eq:Ci_bound} for all epochs.  We do this via an induction argument on the epoch number:
\begin{itemize}[leftmargin=3ex, topsep=0ex] \itemsep0em 
    \item As a base case, recalling that $M_0 = M_{(0)} = D$, we find that \eqref{eq:Ci_LB0} and \eqref{eq:Ci_bound} coincide, with the addition of $\cmax$ arising since once \eqref{eq:Ci_LB0} is exceeded, it may be exceeded by any amount up to $\cmax$.
    \item Fix an epoch number $i > 1$, and suppose that for all $i' < i$, the cost incurred in epoch $i'$ was at most $C^*\big( \frac{\eta_{(i')}}{\beta_{(i')}^{1/2}}, \Mbar_{(i'-1)} \big) \log \frac{|\Mbar_{(i'-1)}| \beta_{(i')}}{ \deltabar^2 \eta_{(i')}^2 } + \cmax$.  By the choice of $\beta_{(i')}$ in \eqref{eq:beta_i}, we find that under the event in Lemma \ref{lem:conf_bounds}, the confidence bounds $\mu_t \pm \sqrt{\beta_{(i')}}\sigma_t$ must have been valid for all $t$ in the epochs $i' < i$, and hence $\Mbar_{(i-1)} \subseteq M_{t_{(i)-1}}$ (\emph{cf.}, \eqref{eq:Mbar_opt}--\eqref{eq:Mbar_lvl}).  
    
    From this, we claim that an analogous argument to the base case implies that \eqref{eq:Ci_LB0} is upper bounded by the right-hand side of \eqref{eq:Ci_bound}, as required.  The only additional argument compared to the base case is noting that $c(S_{t_{(i)}}^*)$ defines the minimum cost to uniformly shrink the posterior standard deviation within $M_{t_{(i)}-1}$ down to $\frac{\eta_{(i)}^2}{\beta_{(i)}}$ after already having chosen $x_1,\dotsc,x_{t_{(i)}-1}$, whereas $C^*\big(\frac{\eta_{(i)}}{\beta_{(i)}^{1/2}}, \Mbar_{(i-1)}\big)$ is defined analogously for the set $\Mbar_{(i-1)}$ with no previously-chosen points.  The latter clearly upper bounds the former.
\end{itemize}

Finally, we check the conditions for $\epsilon$-accuracy in Definition \ref{def:eps_acc}.  In the case of BO, summing \eqref{eq:Ci_bound} over all of the epochs such that $4(1+\deltabar)\eta_{(i-1)} > \epsilon$ yields \eqref{eq:C_eps}; recall from \eqref{eq:Mbar_opt} that after any epoch $i$ such that $4(1+\deltabar)\eta_{(i)} \le \epsilon$, all points are at most $\epsilon$-suboptimal.  We also note that all true maxima must remain in $M_t$, due to the fact that we showed $\beta_{(i)}$ yields valid confidence bounds with high probability, and we only ever discard points that are deemed suboptimal according to those bounds.  For LSE, a similar conclusion follows from \eqref{eq:Mbar_lvl} by summing over all epochs such that $2(1+\deltabar)\eta_{(i-1)} > \frac{\epsilon}{2}$, which is the same as $4(1+\deltabar)\eta_{(i-1)} > \epsilon$.  Once again, all points in $H_t$ and $L_t$ are correct due to the validity of our confidence bounds.

\section{Simplified Result for the Homoscedastic and Unit-Cost Setting} \label{sec:proof_simp}

Since we are focusing on unit costs $c(x) = 1$, the cost simply corresponds to the number of rounds $T$. To highlight this fact, we replace $C^*$ in \eqref{eq:C*} by
\begin{equation}
    T^*(\xi,M) = \min_S \Big\{ |S| \,:\, \max_{\xbar \in M}\sigma_{0|S}(\xbar) \le  \xi \Big\}, \label{eq:T*}
\end{equation}
and similarly replace \eqref{eq:Ti_bound}--\eqref{eq:T_sum} by
\begin{gather}
    T_{(i)} \ge T^*\bigg(\frac{\eta_{(i)}}{\beta_{(i)}^{1/2}}, \Mbar_{(i-1)}\bigg) \log\frac{ |\Mbar_{(i-1)}| \beta_{(i)} }{ \deltabar^2 \eta_{(i)}^2 } + 1 \label{eq:Ti_bound} \\
    \beta_{(i)} \ge 2\log\frac{ |\Mbar_{(i-1)}| \big(\sum_{i' \le i} T_{(i')}\big)^2 \pi^2 }{ 6\delta } \label{eq:betai_T} \\
    T_{\epsilon} = \sum_{i \,:\, 4(1+\deltabar)\eta_{(i-1)} > \epsilon} T_{(i)}. \label{eq:T_sum}
\end{gather}

In this section, we prove the following as an application of Theorem \ref{thm:general}.

\begin{corr} \label{thm:standard}
     Fix $\epsilon >0$ and $\delta \in (0,1)$, define $\beta_T = 2\log\frac{|D| T^2 \pi^2}{ 6\delta }$, and set $\eta_{(1)} = 1$ and $r = \frac{1}{2}$.  There exist choices of $\beta_{(i)}$ (not depending on the time horizon $T$) such that we have $\epsilon$-accuracy with probability at least $1-\delta$ once the following condition holds:
    \begin{align}
        T &\ge  \bigg( C_1 \gamma_T \beta_T \frac{ 96(1+\deltabar)^2  }{ \epsilon^2 } + 2\Big\lceil \log_{2}\frac{8(1+\deltabar) }{\epsilon} \Big\rceil \bigg) \log\frac{ 16(1+\deltabar)^2 |D| \beta_T }{ \deltabar^2 \epsilon^2 }, \label{eq:simplified}
     \end{align}
    where $C_1 = \frac{1}{\log(1+\sigma^{-2})}$.  This condition is of the form $T \ge \Omega^*\big( \frac{C_1 \gamma_T \beta_T}{ \epsilon^2 } + 1 \big)$.
\end{corr}

We bound the cardinality of $S$ in \eqref{eq:T*} by considering a procedure that greedily picks $\argmax_{x \in M} \sigma_{t}(x)$. We claim that after selecting $k$  points according to this procedure to construct a set $S_k$, we have
\begin{equation}
    \max_x \sigma_{0|S_k}^2(x) \le C_1 \frac{\gamma_k}{k}, 
\end{equation}
where $C_1 = \frac{1}{\log(1+\sigma^{-2})}$. This is seen by writing
\begin{align}
    k \max_{x \in M} \sigma_{0|S_k}^2(x) 
        &= k \sigma_{0|S_k}^2(x_k) \\
        &\le \sum_{j=1}^k \sigma_{0|S_j}^2(x_j) \\
        &\le \frac{1}{\log(1+\sigma^{-2})} \gamma_k, \label{eq:gamma_bound}
\end{align} 
where we respectively used that $x_k$ maximizes $\sigma_{0|S_k}$, that $\sigma_{0|S_i}(x_i)$ always decreases as more points are chosen, and the bound on the sum of variances of sampled points from \cite[Lemma 5.4]{Sri12}.

Identifying $k$ with $T^*$, and $\max_{x\in M}\sigma_{0|S_k}(x)$ with $\xi = \frac{\eta}{\beta^{1/2}}$ (for some $\eta$ and $\beta$ to be specified), we obtain from \eqref{eq:gamma_bound} that
\begin{equation}
    T^*\bigg(\frac{\eta}{\beta^{1/2}},M\bigg) \le \min \bigg\{ T^*\,:\, T^* \ge \frac{C_1 \gamma_{T^*} \beta}{\eta^2} \bigg\}. \label{eq:T*_bound}
\end{equation}
Consider the value $T^*\big(\frac{\eta_{(i)}}{\beta_{(i)}^{1/2}},\Mbar_{(i-1)}\big)$ corresponding to the parameters $\eta = \eta_{(i)}$ and $\beta = \beta_{(i)}$ associated with epoch $i$. Letting $T = T_{\epsilon}$ denote the total time horizon, and using \eqref{eq:T_sum}, we find that $\beta_{(i)}$ in \eqref{eq:betai_T} can be upper bounded by $2\log\frac{|D| T^2 \pi^2}{ 6\delta }$, which is precisely $\beta_T$.  By similarly using the monotonicity of $\gamma_t$, we obtain 
\begin{equation}
    T^*\bigg(\frac{\eta_{(i)}}{\beta_{(i)}^{1/2}},\Mbar_{(i-1)}\bigg) \le  \frac{C_1 \gamma_{T} \beta_T}{\eta_{(i)}^2} + 1, \label{eq:T*_bound2}
\end{equation}
where the addition of one is to account for possible rounding up to the nearest integer.

Using \eqref{eq:T*_bound2}, we find that in order for \eqref{eq:Ti_bound} to hold it suffices that
\begin{equation}
     T_{(i)} \ge \bigg( \frac{C_1 \gamma_{T} \beta_T}{\eta_{(i)}^2} + 1 \bigg) \log\frac{ |\Mbar_{(i-1)}| \beta_{(i)} }{ \deltabar^2 \eta_{(i)}^2 } + 1. \label{eq:Ti_bound2}
\end{equation}
Since we are only considering values of $i$ such that $4(1+\deltabar)\eta_{(i-1)} > \epsilon$, and since $\Mbar_{(i-1)} \subseteq D$, we can upper bound the logarithm by $\log\frac{16(1+\deltabar)^2 |D| \beta_{(i)}}{ \deltabar^2 \epsilon^2 } > 1$, and hence in order for \eqref{eq:Ti_bound2} to hold it suffices that
\begin{equation}
     T_{(i)} \ge \bigg( \frac{C_1 \gamma_{T} \beta_T}{\eta_{(i)}^2} + 2 \bigg) \log\frac{16(1+\deltabar)^2 |D| \beta_T}{ \deltabar^2 \epsilon^2 }. \label{eq:Ti_bound3}
\end{equation}
We also note that since $\eta_{(i)} = \eta_{(1)}r^{i-1}$, the condition $4(1+\deltabar)\eta_{(i-1)} > \epsilon$ is equivalent to
\begin{align}
    & 4(1+\deltabar)\eta_{(1)}r^{i-2} > \epsilon \\
    & \iff r^{i-2} > \frac{\epsilon}{4(1+\deltabar)\eta_{(1)}} \\
    & \iff i < 2 + \log_{1/r}\frac{4(1+\deltabar)\eta_{(1)}}{\epsilon} \\
    & \iff i \le \Big\lceil \log_{1/r}\frac{4(1+\deltabar)\eta_{(1)}}{\epsilon} \Big\rceil + 1 \\
    & \iff i \le \Big\lceil \log_{1/r}\frac{4(1+\deltabar)\eta_{(1)}}{r\epsilon} \Big\rceil, \label{eq:equiv_i}
\end{align}
where in the last line we used $\log_{1/r}\frac{1}{r} = 1$.  Summing \eqref{eq:Ti_bound3} over all such $i$ in accordance with \eqref{eq:T_sum}, we obtain following sufficient condition on the time horizon for $\epsilon$-accuracy:
\begin{equation}
    T \ge  \bigg( C_1 \gamma_T \beta_T \sum_{i=1}^{\lceil\log_{1/r}\frac{4(1+\deltabar)\eta_{(1)}}{\epsilon}\rceil + 1} \frac{1}{\eta_{(i)}^2} + 2\Big\lceil \log_{1/r}\frac{4(1+\deltabar) \eta_{(1)}}{r\epsilon} \Big\rceil \bigg) \log\frac{ 16(1+\deltabar)^2 |D| \beta_T }{ \deltabar^2 \epsilon^2 }. \label{eq:Tbound_sum}
\end{equation}
Finally, we weaken this condition by upper bounding the summation as follows:
\begin{align}
    \sum_{i=1}^{\lceil\log_{1/r}\frac{4(1+\deltabar)\eta_{(1)}}{\epsilon}\rceil + 1} \frac{1}{\eta_{(i)}^2} 
        &= \sum_{i=1}^{\lceil\log_{1/r}\frac{4(1+\deltabar)\eta_{(1)}}{\epsilon}\rceil + 1} \frac{1}{\eta_{(1)}^2 r^{2(i-1)} } \\
        &= \sum_{i=0}^{\lceil\log_{1/r}\frac{4(1+\deltabar)\eta_{(1)}}{\epsilon}\rceil } \frac{1}{\eta_{(1)}^2 r^{2i} } \\
        &\le \frac{1}{r^2(1-r^2)} \frac{ 16(1+\deltabar)^2  }{ \epsilon^2 }, \label{eq:eval_sum}
\end{align}
where the last line follows from the identity $\sum_{i=0}^{\lceil\log_{1/r}A \rceil } \frac{1}{r^{2i}} \le \frac{1}{r^2(1-r^2)} A^2 $ for $\log_{1/r}A > 0$.  Substituting $r=\frac{1}{2}$ and $\eta_{(1)} = 1$ concludes the proof; the former yields $\frac{1}{r^2(1-r^2)} = \frac{16}{3} \le 6$.

\section{Proof of Improved Noise Dependence (Corollary \ref{thm:noise_dep}))} \label{sec:proof_noise}

The bound in \eqref{eq:gamma_bound} is based on the inequality \cite[Lemma 5.4]{Sri12}
\begin{equation}
    \frac{\sigma_t^2}{\sigma^2} \le C_1 \log\bigg(1+\frac{\sigma_t^2}{\sigma^2}\bigg) \label{eq:gamma_bound1}
\end{equation}
for $\sigma_t^2 \in [0,1]$ (with $C_1 = \frac{\sigma^{-2}}{\log(1+\sigma^{-2})}$), which can be very loose when $\sigma^2$ is small.  Our starting point to improve the noise dependence is to note that the following holds under the more restrictive condition $\sigma_t^2 \le \sigma^2$:
\begin{align}
    \sigma_t^2 
        &= \sigma^2 \frac{\sigma_t^2}{\sigma^2} \\
        &\le 2\sigma^2\log \bigg( 1+\frac{\sigma_t^2}{\sigma^2} \bigg), \label{eq:gamma_bound2}
\end{align}
where we have used the fact that $\alpha \le 2\log(1+\alpha)$ for $\alpha \in [0,1]$.

The idea now is to use \eqref{eq:gamma_bound2} in the epochs that are late enough so that $\sigma_t^2 \le \sigma^2$, and \eqref{eq:gamma_bound} in the earlier epochs.  Since $(1+\deltabar)\eta_{(i)}$ is the confidence level obtained after epoch $i$, and since $\beta_{(i)}^{1/2}\sigma_t$ is the confidence level after time $t$, we find that in order to ensure $\sigma_t^2 \le \sigma^2$ it suffices that $\frac{(1+\deltabar)^2\eta_{(i)}^2}{\beta_{(i)}} \le \sigma^2$.  Moreover, our choice of $\beta_{(i)}$ in \eqref{eq:beta_i} is always greater than one when $|D| \ge 2$ (which is a trivial condition), and hence we can weaken this condition to $(1+\deltabar)^2\eta_{(i)}^2 \le \sigma^2$, and write
\begin{equation}
    \sum_{i} T_{(i)} \le \sum_{i\,:\, (1+\deltabar)^2\eta_{(i-1)}^2 > \sigma^2} T_{(i)}^{(C_1)} + \sum_{i} T_{(i)}^{(2\sigma^2)},
\end{equation}  
where $T_{(i)}^{(C_1)}$ denotes bound on $T_{(i)}$ in \eqref{eq:Ti_bound3} based on \eqref{eq:gamma_bound}, and $T_{(i)}^{(C_1)}$ denotes the analogous bound based on \eqref{eq:gamma_bound2} with $2\sigma^2$ in place of $C_1$.  Similarly to \eqref{eq:equiv_i}, the first summation is over a subset of the range $i \le \lceil \log_{1/r}\frac{(1+\deltabar)\eta_{(1)}}{r\sigma} \rceil$, and it follows that the condition \eqref{eq:Tbound_sum} may be replaced by
\begin{multline}
    T \ge  \bigg( 2\sigma^2 \gamma_T \beta_T \sum_{i=1}^{\lceil\log_{1/r}\frac{4(1+\deltabar)\eta_{(1)}}{\epsilon}\rceil + 1} \frac{1}{\eta_{(i)}^2} + 2\Big\lceil \log_{1/r}\frac{4(1+\deltabar) \eta_{(1)}}{r\epsilon} \Big\rceil \bigg) \log\frac{ 16(1+\deltabar)^2 |D| \beta_T }{ \deltabar^2 \epsilon^2 } \\
    + \bigg( C_1 \gamma_T \beta_T \sum_{i=1}^{\lceil\log_{1/r}\frac{(1+\deltabar)\eta_{(1)}}{\sigma}\rceil + 1} + 2\Big\lceil \log_{1/r}\frac{(1+\deltabar)\eta_{(1)}}{r\sigma} \Big\rceil \bigg) \log\frac{ 16(1+\deltabar)^2 |D| \beta_T }{ \deltabar^2 \epsilon^2 }. \label{eq:Tbound_sum2}
\end{multline}
The first summation is handled in the same way as the previous subsection, and the second summation is upper bounded by writing
\begin{align}
    \sum_{i=1}^{\lceil\log_{1/r}\frac{(1+\deltabar)\eta_{(1)}}{\sigma}\rceil + 1} \frac{1}{\eta_{(i)}^2} 
        &= \sum_{i = 1}^{\lceil \log_{1/r} \frac{(1+\deltabar)\eta_{(1)}}{\sigma} \rceil + 1 } \frac{1}{\eta_{(1)}^2 r^{2(i-1)} } \label{eq:noise_dep3} \\
        & \le \frac{(1+\deltabar)^2}{r^2(1-r^2)} \frac{ 1 }{ \sigma^2 },  \label{eq:noise_dep4}
\end{align}
where \eqref{eq:noise_dep3} follows since $\eta_{(i)} = \eta_{(1)} r^{i-1}$, and \eqref{eq:noise_dep4} follows in the same way as \eqref{eq:eval_sum}.  Once again, setting $r=\frac{1}{2}$ and $\eta_{(1)} = 1$ concludes the proof, with the third term in \eqref{eq:improved} coming from the identity $2\big\lceil \log_{2}\frac{8(1+\deltabar)}{\epsilon}\big\rceil + 2\big\lceil \log_{2}\frac{2(1+\deltabar)}{\sigma} \big\rceil \le 2\big\lceil \log_{2}\frac{32(1+\deltabar)^2 }{\epsilon\sigma} \big\rceil $.

\section{Proof for the Setting of Choosing Noise (Corollary \ref{thm:choose_noise})} \label{sec:proof_choose}

The proof follows the same arguments as those of Appendices \ref{sec:proof_simp} and \ref{sec:proof_noise}, with $C^*$ being upper bounded in $K$ different ways, one for each possible noise level.  The choice $\beta_T = 2\log\frac{|D| T^2 \cmax^2 \pi^2}{ 6\delta\cmin^2 }$ arises as a simple upper bound to the right-hand side of \eqref{eq:beta_i} resulting from the fact that $\sum_{t=1}^T c(x_t) \le \cmax T$. 

\end{document}